\documentclass{article}

\usepackage[final]{neurips_2026}

\usepackage[T1]{fontenc}
\usepackage{hyperref}
\usepackage{url}
\usepackage{booktabs}
\usepackage{amsfonts}
\usepackage{amsmath}
\usepackage{amssymb}
\usepackage{nicefrac}
\usepackage{microtype}
\usepackage{graphicx}
\usepackage{xcolor}
\usepackage{algorithm}
\usepackage{algorithmic}
\usepackage{subcaption}
\usepackage{multirow}
\usepackage{wrapfig}
\usepackage{colortbl}

\graphicspath{{./}}

\title{The Spectral Lifecycle of Transformer Training: \\
Transient Compression Waves, Persistent Spectral Gradients, \\
and the Q/K--V Asymmetry}

\author{
  Yi Liu \\
  \texttt{lylewis@outlook.com}
}

\begin{document}

\maketitle

\begin{abstract}
We present the first systematic study of weight matrix singular value spectra \emph{during} transformer pretraining, tracking full SVD decompositions of every weight matrix at 25-step intervals across three model scales (30M--285M parameters).
We discover three phenomena:
\textbf{(1)~Transient Compression Waves:} stable rank compression propagates as a traveling wave from early to late layers, creating a dramatic gradient that peaks early then \emph{reverses}---late layers eventually over-compress past early layers.
\textbf{(2)~Persistent Spectral Gradients:} the power-law exponent~$\alpha$ develops a permanent depth gradient forming a non-monotonic inverted-U in deeper models, with peaks shifting toward earlier layers as depth increases.
\textbf{(3)~Q/K--V Functional Asymmetry:} value/output projections compress uniformly while query/key projections carry the full depth-dependent dynamics.
The dissociation between transient compression and persistent spectral shape reveals that \emph{rank and spectral shape encode fundamentally different information about training}.
We formalize this as a two-timescale dynamical model and derive scaling laws ($\Delta\alpha \propto L^{0.26}$, $R^2{=}0.99$).
We validate on nine models across three families (custom, GPT-2, Pythia; 30M--1B parameters; 8--36 layers), demonstrate that $\alpha$ predicts layer importance ($\rho{=}0.69$--$0.84$, $p{<}0.02$), and show that spectral-guided pruning outperforms Last-N heuristics by $1.1{\times}$--$3.6{\times}$ across seven models in two families (GPT-2 124M--774M, Pythia 160M--1B), with worst-vs-best gaps up to $23.7{\times}$ confirming the causal role of spectral structure.
\end{abstract}

\section{Introduction}
\label{sec:intro}

Training large language models costs millions of dollars, yet we understand surprisingly little about what happens inside them during training. Despite advances in scaling laws~\citep{kaplan2020scaling, hoffmann2022training}, architectures~\citep{gu2024mamba}, and training recipes~\citep{touvron2023llama}, practitioners monitor loss curves---a single scalar---hoping that billions of parameters evolve sensibly.

Random matrix theory (RMT) offers a richer lens. The Heavy-Tail Self-Regularization (HT-SR) framework~\citep{martin2021implicit, martin2021htsr_jmlr} showed that \emph{trained} networks exhibit heavy-tailed singular value distributions correlating with generalization, and recent work has exploited this for pruning~\citep{lu2024alphapruning} and studied per-matrix spectral dynamics~\citep{yunis2024spectral}. However, the critical question of how spectral structure \emph{varies across layers and evolves over training}---the spatiotemporal picture---remains unexplored. Existing work either examines static snapshots or tracks individual matrices without studying inter-layer relationships.

We conduct the first comprehensive \textbf{spectral time-lapse} of transformer pretraining, tracking full SVDs of every weight matrix at fine-grained intervals across multiple scales. This yields a spatiotemporal dataset of over 150{,}000 SVD snapshots that reveals phenomena invisible to scalar monitoring.

Our central discovery is a \textbf{dissociation between compression and spectral shape}: the \emph{compression} (stable rank reduction) follows a transient wave that eventually equilibrates, while the \emph{spectral shape} (power-law exponent~$\alpha$) develops a permanent gradient. Compression measures \emph{how much} structure has emerged; $\alpha$ measures \emph{what kind}.

\paragraph{Contributions.} \textbf{(1)} We discover transient compression waves propagating from early to late layers at ${\sim}$80--100 steps/layer, with a gradient that reverses from negative to positive as late layers over-compress (Section~\ref{sec:transient}). \textbf{(2)} We identify persistent spectral gradients where $\alpha$ forms a non-monotonic inverted-U that strengthens throughout training, with the peak shifting toward earlier relative depth in deeper models (Section~\ref{sec:persistent}). \textbf{(3)} We reveal Q/K--V functional asymmetry: V/O projections compress uniformly while Q/K carry depth-dependent dynamics (Section~\ref{sec:qkv}). \textbf{(4)} We formalize a two-timescale dynamical model with scaling laws (Section~\ref{sec:theory}). \textbf{(5)} We validate on pretrained GPT-2 (124M--774M) and Pythia (160M--1B) models with training checkpoints confirming the theory (Section~\ref{sec:validation}). \textbf{(6)} We show $\alpha$ predicts layer importance and enables spectral-guided pruning that outperforms Last-N by $1.1$--$3.6\times$ across seven models in two families: GPT-2 (124M--774M) and Pythia (160M--1B), holding from 12 to 36 layers (Section~\ref{sec:applications}).

\section{Related Work}
\label{sec:related}

\paragraph{RMT and Spectral Analysis.} The HT-SR framework~\citep{martin2021implicit, martin2021htsr_jmlr} established that trained networks exhibit heavy-tailed spectra correlating with generalization. \citet{lu2024alphapruning} leveraged $\alpha$ for layerwise pruning \emph{ratios} in pretrained LLMs, showing per-layer $\alpha$ variation predicts prunability. We complement this by revealing \emph{how} $\alpha$ gradients emerge during training and showing they predict layer \emph{importance}, not just prunability. \citet{staats2024small} showed small singular values carry meaning. \citet{huang2025early} used spectral density evolution for early stopping. All analyze static or coarse-grained snapshots; we track fine-grained dynamics.

\paragraph{Spectral Dynamics of Weights.} Most closely related, \citet{yunis2024spectral} study singular value and vector dynamics during optimization, identifying a ``bulk-plus-spike'' pattern where a few top singular values grow while the bulk remains near initialization. Our work differs in three ways: (i)~we analyze \emph{inter-layer} gradients and their temporal evolution, not individual-matrix dynamics; (ii)~we discover the transient/persistent dissociation between compression and spectral shape---a phenomenon invisible when studying single matrices; (iii)~we validate across 9 models in 3 families including training checkpoints. \citet{xu2026spectral_edge} study spectral edge dynamics of parameter \emph{updates}; \citet{olsen2025sgd_spectra} develop an SDE framework for singular value evolution; \citet{spectral_dimension2025} study \emph{activation} spectra. Our analysis of weight spectra across depth reveals complementary spatiotemporal phenomena.

\paragraph{Training Dynamics.} Grokking~\citep{power2022grokking, grokking2026}, the break-even point~\citep{jastrzebski2020break}, edge of stability~\citep{cohen2021gradient}, and double descent~\citep{nakkiran2021deep} reveal phase transitions. Our spectral perspective uncovers a novel transient/persistent dissociation invisible to scalar metrics.

\paragraph{Per-Layer Analysis.} Layer roles have been studied through probing~\citep{tenney2019bert}, knowledge storage~\citep{geva2022transformer}, induction heads~\citep{olsson2022context}, and layer pruning~\citep{men2024shortgpt}. Our spectral metrics capture layer specialization from weights alone, without input data.

\section{Method}
\label{sec:method}

\subsection{Spectral Metrics}

For each weight matrix $\mathbf{W} \in \mathbb{R}^{m \times n}$ with singular values $\sigma_1 \geq \cdots \geq \sigma_k$ ($k = \min(m,n)$), we track:

\textbf{Stable Rank:} $R_s(\mathbf{W}) = \|\mathbf{W}\|_F^2 / \|\mathbf{W}\|_2^2 = \sum_i \sigma_i^2 / \sigma_1^2$, measuring effective dimensionality.

\textbf{Weighted Alpha:} $\log \sigma_i \approx -\alpha \log i + c$ for $i = 1, \ldots, \lfloor 0.2k \rfloor$, measuring power-law tail heaviness.

\textbf{Spectral Entropy:} $H = -(\log_2 k)^{-1} \sum_i p_i \log_2 p_i$, $p_i = \sigma_i^2/\|\boldsymbol{\sigma}\|_2^2$, an independent confirmation metric.

The \textbf{compression onset} for layer $l$ is the first step where $R_s$ drops below $0.9 R_s^{(l)}(0)$. Fitting $t_{\text{onset}}^{(l)} \approx v \cdot l + t_0$ yields the compression wave velocity $v$.

\subsection{Experimental Setup}

We train GPT-2-style transformers at three scales (Table~\ref{tab:models}) on ClimbMix-400B~\citep{karpathy2024} with AdamW ($\beta_1{=}0.9$, $\beta_2{=}0.95$, weight decay 0.1, cosine LR $6{\times}10^{-4}$) for up to 10{,}000 steps. Full SVDs are computed every 25 steps (D8/D12) or 50 steps (D16).

\begin{table}[t]
\centering
\caption{Model configurations.}
\label{tab:models}
\vspace{-3pt}
\small
\begin{tabular}{lccccc}
\toprule
Model & $d_{\text{model}}$ & Layers & Heads & Params & SVD Snapshots \\
\midrule
D8  & 512  & 8  & 8  & 30.4M  & 20{,}050 \\
D12 & 768  & 12 & 12 & 92.8M  & 29{,}674 \\
D16 & 1024 & 16 & 16 & 285.2M & 35{,}142 \\
\bottomrule
\end{tabular}
\vspace{-5pt}
\end{table}

\section{Results}
\label{sec:results}

\subsection{Finding 1: Transient Compression Waves}
\label{sec:transient}

\begin{figure}[t]
    \centering
    \begin{subfigure}[t]{0.48\linewidth}
        \includegraphics[width=\linewidth]{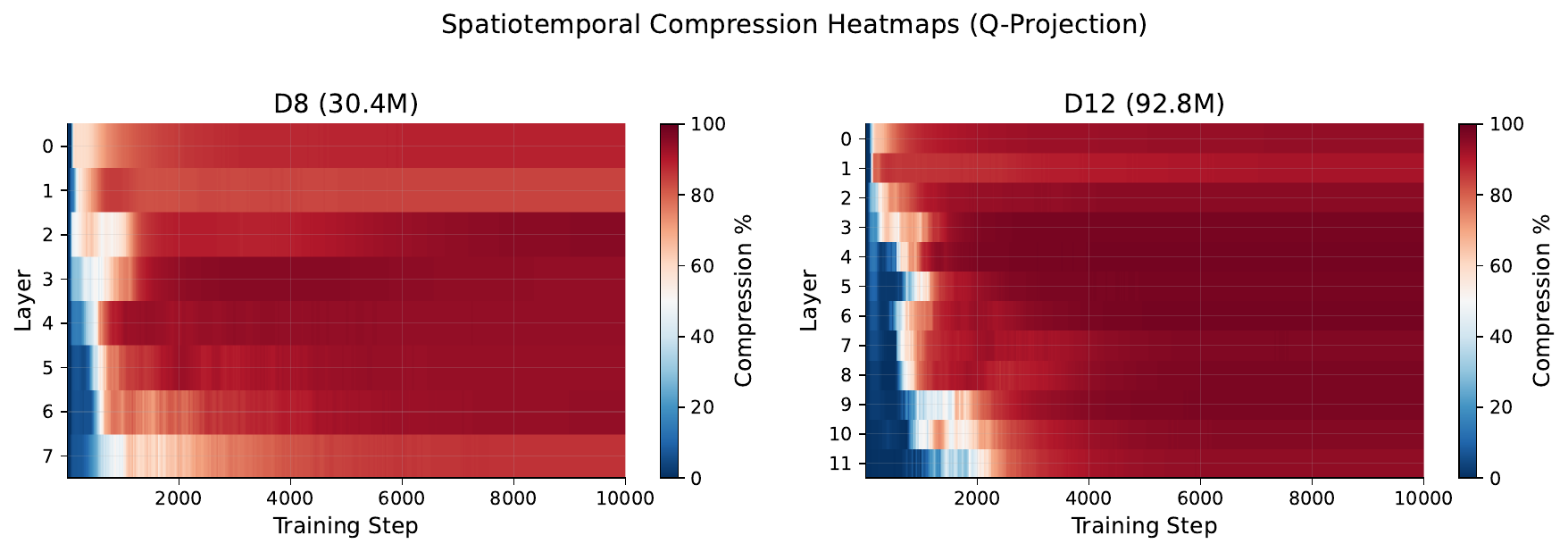}
        \caption{Compression heatmaps (D8, D12).}
    \end{subfigure}
    \hfill
    \begin{subfigure}[t]{0.48\linewidth}
        \includegraphics[width=\linewidth]{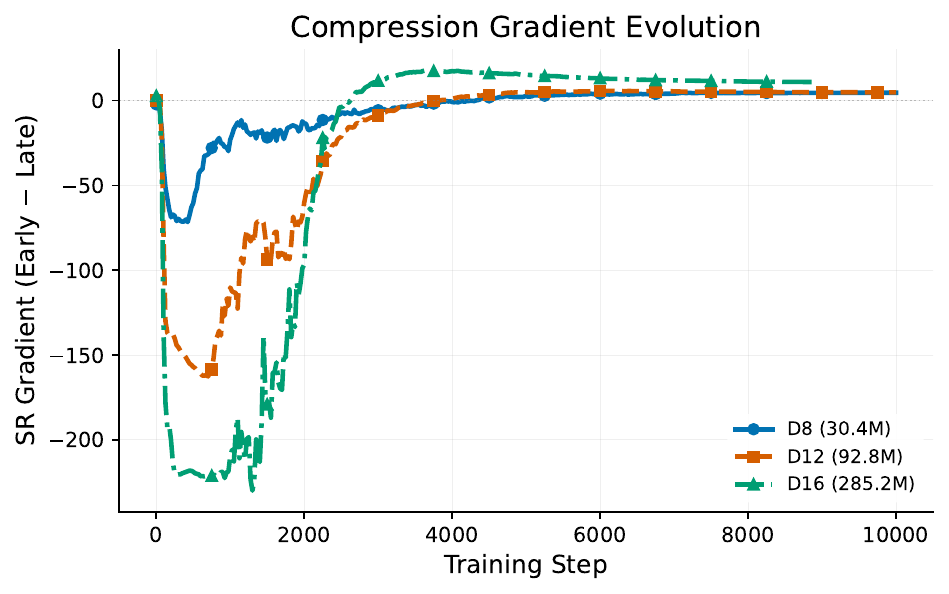}
        \caption{Gradient evolution (all scales).}
    \end{subfigure}
    \caption{\textbf{Transient compression waves.} (a)~Spatiotemporal heatmaps show the diagonal wave front. (b)~The compression gradient transitions from negative to positive in all models, with D16 showing the largest magnitude swings.}
    \label{fig:compression}
    \vspace{-5pt}
\end{figure}

Stable rank compression propagates as a \textbf{traveling wave} through the network. Early layers (L0--L1) compress rapidly in the first 500 steps, while late layers lag by hundreds of steps at ${\sim}$80--100 steps per layer (Figure~\ref{fig:compression}).

Crucially, the inter-layer gradient is \textbf{not permanent}. Figure~\ref{fig:compression}(b) tracks the compression gradient (L$_\text{first} - $ L$_\text{last}$ SR) across training. All models transition from negative (early layers compress first) to positive (late layers over-compress). In D16, the gradient swings from $-59.6$ at step 500 to $+18.8$ by step 5{,}000, with deep layers reaching SR$\,{\approx}\,$4.3 while early layers stabilize at SR$\,{\approx}\,$23.

\begin{table}[t]
\centering
\caption{Compression gradient evolution (L$_\text{first} - $ L$_\text{last}$ average SR). All models show a negative$\to$positive transition.}
\label{tab:gradient}
\vspace{-3pt}
\small
\begin{tabular}{lcccccc}
\toprule
& Step 250 & Step 500 & Step 1K & Step 2K & Step 5K & Final \\
\midrule
D8   & $-$23.1 & $-$15.0 & +2.9 & +14.4 & +19.7 & +22.1 \\
D12  & $-$42.2 & $-$25.9 & $-$2.6 & +8.1 & +17.0 & +17.2 \\
D16  & $-$41.7 & $-$59.6 & $-$19.5 & $-$2.6 & +17.8 & +18.8 \\
\bottomrule
\end{tabular}
\vspace{-5pt}
\end{table}

\subsection{Finding 2: Persistent Spectral Shape Gradients}
\label{sec:persistent}

\begin{figure}[t]
    \centering
    \begin{subfigure}[t]{0.48\linewidth}
        \includegraphics[width=\linewidth]{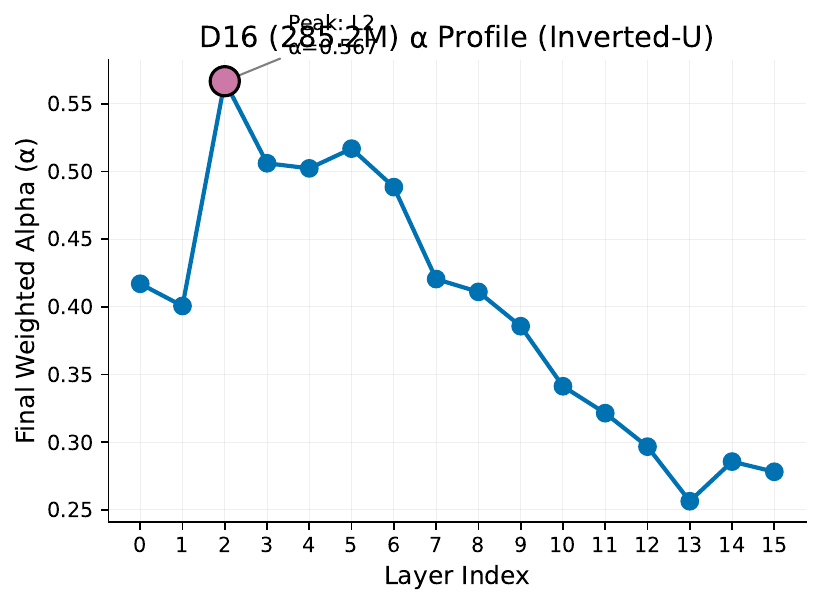}
        \caption{D16 inverted-U $\alpha$ profile.}
    \end{subfigure}
    \hfill
    \begin{subfigure}[t]{0.48\linewidth}
        \includegraphics[width=\linewidth]{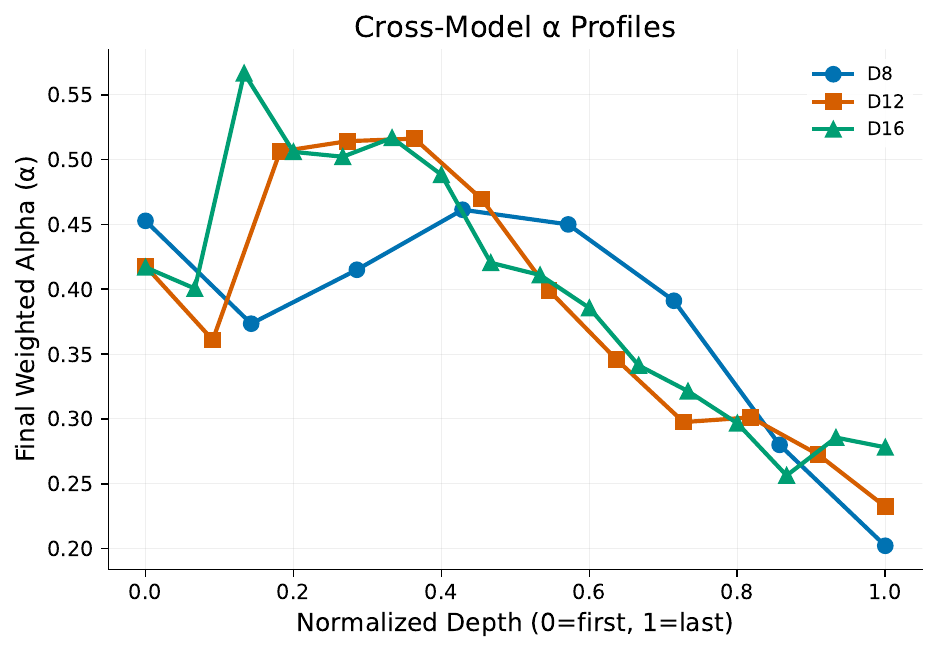}
        \caption{Cross-model comparison.}
    \end{subfigure}
    \caption{\textbf{Persistent spectral gradients.} (a)~D16 Q-projection $\alpha$ across layers showing the inverted-U. (b)~Cross-model comparison on normalized depth: the peak shifts earlier in deeper models.}
    \label{fig:alpha_profiles}
    \vspace{-5pt}
\end{figure}

In stark contrast to compression, the power-law exponent $\alpha$ develops a \emph{permanent} depth gradient that \emph{strengthens} throughout training (Figure~\ref{fig:alpha_profiles}). In deeper models, this gradient is non-monotonic---an \textbf{inverted-U} peaking at early-middle layers.

In D16, $\alpha$ peaks at L2 ($\alpha = 0.567$) and drops to $\alpha = 0.256$ at L13---a 121\% spread. The peak position shifts toward earlier \emph{relative} depth in deeper models: 43\% (D8), 33\% (D12), 13\% (D16), suggesting the heavy-tail zone occupies a fixed number of layers (${\sim}$2--4) regardless of total depth. Spectral entropy independently confirms this gradient (Appendix~\ref{app:entropy}).

\begin{figure}[t]
    \centering
    \begin{subfigure}[t]{0.48\linewidth}
        \includegraphics[width=\linewidth]{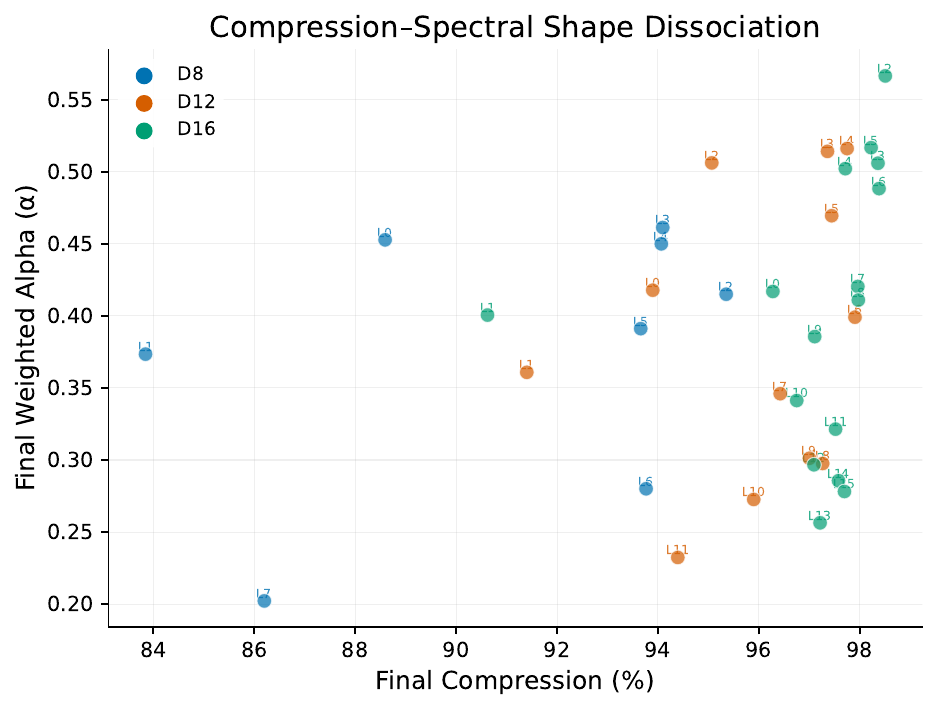}
        \caption{Compression vs.\ $\alpha$ (each point = one layer).}
    \end{subfigure}
    \hfill
    \begin{subfigure}[t]{0.48\linewidth}
        \includegraphics[width=\linewidth]{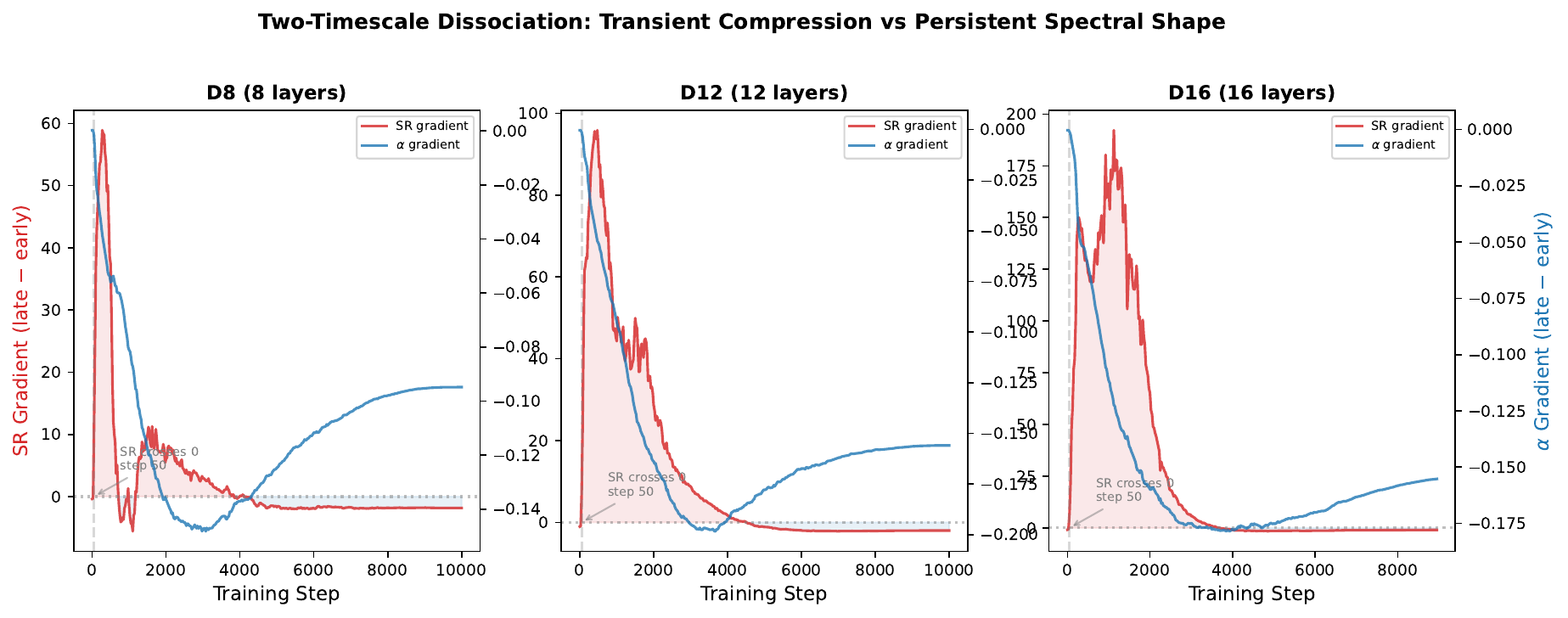}
        \caption{SR gradient (transient) vs.\ $\alpha$ gradient (persistent).}
    \end{subfigure}
    \caption{\textbf{The two-timescale dissociation.} (a)~Layers reach similar compression but divergent $\alpha$. (b)~SR gradient reverses while $\alpha$ gradient monotonically strengthens---the hallmark of two-timescale dynamics.}
    \label{fig:dissociation}
    \vspace{-5pt}
\end{figure}

\paragraph{The Dissociation.} Figure~\ref{fig:dissociation}(a) plots final compression vs.\ final $\alpha$ for each layer. Despite layers converging to similar compression levels, their $\alpha$ values remain widely spread---\emph{rank and spectral shape encode fundamentally different information}. Figure~\ref{fig:dissociation}(b) directly visualizes the two timescales: the SR gradient is transient (reverses sign) while the $\alpha$ gradient is persistent (monotonically strengthens).

\subsection{Finding 3: Q/K--V Functional Asymmetry}
\label{sec:qkv}

Within each attention layer, Q and K projections exhibit the full depth-dependent spectral dynamics (compression wave, $\alpha$ gradient), while V and O projections compress uniformly to $>$90\% regardless of depth. This asymmetry suggests that \emph{attention selection is the adaptive computation} while value transformation is more generic. Full results are in Appendix~\ref{app:qkv}; multi-seed reproducibility ($n{=}4$) is confirmed in Appendix~\ref{app:reproducibility}.

\section{A Two-Timescale Theory of Spectral Evolution}
\label{sec:theory}

\subsection{Dynamical Model}

We model the spectral state of layer $l$ at step $t$ as:
\begin{align}
    \frac{dR_s^{(l)}}{dt} &= -\lambda_R \cdot \phi(l, t) \cdot \bigl(R_s^{(l)} - R_s^*\bigr) + \xi_R^{(l)}(t) \label{eq:sr_dynamics} \\
    \frac{d\alpha^{(l)}}{dt} &= \lambda_\alpha \cdot \psi(l) \cdot \bigl(\alpha^*(l) - \alpha^{(l)}\bigr) + \xi_\alpha^{(l)}(t) \label{eq:alpha_dynamics}
\end{align}
where $\lambda_R \gg \lambda_\alpha$ encodes the timescale separation. The compression driving function $\phi(l, t)$ depends on both layer and time (producing the traveling wave: early layers receive structured input immediately, late layers must wait). The shape function $\psi(l)$ depends primarily on layer position. The shared equilibrium $R_s^*$ is approximately constant (same information bottleneck), while $\alpha^*(l)$ varies by layer (different computational roles).

This model predicts: (i)~SR gradients are transient (all converge to $R_s^*$); (ii)~$\alpha$ gradients are persistent ($\alpha^*(l)$ varies); (iii)~the compression wave propagates forward (input-driven $\phi$).

\subsection{Scaling Laws}

\begin{wraptable}{r}{0.42\textwidth}
\vspace{-12pt}
\centering
\caption{Scaling laws.}
\label{tab:scaling}
\vspace{-3pt}
\small
\begin{tabular}{lccc}
\toprule
& D8 & D12 & D16 \\
\midrule
$\alpha_{\max}$ & 0.461 & 0.516 & 0.567 \\
$\Delta\alpha$ & 0.259 & 0.284 & 0.310 \\
Peak $l^*/L$ & 0.43 & 0.36 & 0.13 \\
$v_{\text{wave}}$ & 102 & 131 & 142 \\
\bottomrule
\end{tabular}
\vspace{-10pt}
\end{wraptable}

Our experiments reveal precise scaling (Table~\ref{tab:scaling}):
\begin{align}
    \Delta\alpha &\propto L^{0.26}, \;\; R^2{=}0.99 \label{eq:spread} \\
    \alpha_{\max} &\propto L^{0.30}, \;\; R^2{=}1.00 \label{eq:peak} \\
    l^*/L &\approx {-}0.037L + 0.75, \;\; R^2{=}0.91 \label{eq:position}
\end{align}
The sublinear growth of $\Delta\alpha$ means each depth doubling increases differentiation by only ${\sim}$20\%. The linear decrease of $l^*/L$ confirms that the heavy-tail zone occupies a fixed number of layers.

\paragraph{Gradient Flow Interpretation.} The outer product structure $\Delta\mathbf{W}^{(l)} \propto \boldsymbol{\delta}^{(l)} \cdot (\mathbf{h}^{(l-1)})^\top$ explains both timescales~\citep{pennington2018spectrum}: \emph{compression converges} because total information per layer is bounded by data entropy---all layers process the same sequences through the same loss, converging to a shared effective rank. \emph{$\alpha$ diverges} because the spectral structure of gradient signals differs by position: early layers receive gradients refined through all subsequent layers (heavier tails, fewer dominant directions), while late layers receive more diffuse signals closer to the raw loss.

\paragraph{The Inverted-U Pattern.} L0--L1 have slightly lower $\alpha$ because they perform low-level embedding processing~\citep{tenney2019bert}. Early-middle layers (L2--L5) perform the most complex transformations---abstracting from tokens to semantics~\citep{geva2022transformer}---and develop the heaviest tails. Late layers handle increasingly specialized but narrower computations. The peak shifting toward earlier relative depth in deeper models (Eq.~\ref{eq:position}) confirms that the heavy-tail zone occupies a fixed number of layers. This connects to neural collapse~\citep{papyan2020prevalence}: deep layers approach collapse-like configurations (low $\alpha$) while early layers maintain broader spectral support.

\section{Validation on Pretrained Models}
\label{sec:validation}

\subsection{GPT-2 Family (124M--774M)}

We analyze GPT-2 Small (12L), Medium (24L), and Large (36L)~\citep{radford2019language}---trained on ${\sim}$40B tokens, a fundamentally different regime from our experiments. All three core phenomena persist:

\textbf{Persistent $\alpha$ gradients:} Q-$\alpha$ peak shifts toward earlier depth in deeper models (L11 in Small, L0 in Medium, L1 in Large), consistent with our D8$\to$D12$\to$D16 trend.

\textbf{Amplified Q/K--V asymmetry:} V-$\alpha$ drops 40--55\% from early to late layers---more extreme than in our partially-trained models.

\textbf{Equilibrated compression:} SR gradients are weak, consistent with full convergence of the fast timescale.

\subsection{Pythia Suite (160M--1B): Temporal Validation}

The Pythia models~\citep{biderman2023pythia} provide training checkpoints (steps 0--143K), enabling direct temporal validation:

\textbf{Compression waves confirmed:} At step 1K, early layers have compressed (SR$\,{\approx}\,$22--99) while late layers remain near initialization. By step 143K, late layers are maximally compressed (SR$\,{<}\,$3).

\textbf{$\alpha$ gradient strengthens monotonically:} Pythia-160M $\alpha$ spread: $0.003 \to 0.058 \to 0.137 \to 0.282 \to 0.333$ across training---never reversing, confirming persistent gradients.

\textbf{Peak migration:} In Pythia-410M (24L), the $\alpha$ peak migrates from L10 (step 1K) to L22 (step 143K), and a sharp phase transition emerges at L13--L14 where late layers enter an extreme spectral regime (Q-SR$\,{<}\,$3, $\alpha > 0.34$). This migration reveals that peak position depends on training duration.

\textbf{Billion-scale confirmation:} Pythia-1B (16L, 1B params) confirms $\alpha$ gradient emergence from flat initialization, with the peak migrating from L7 (step 1K) to L15 (step 5K) to L3 (step 143K)---the same late$\to$early migration seen in GPT-2.

\subsection{Cross-Family Synthesis}

\begin{figure}[t]
    \centering
    \includegraphics[width=\linewidth]{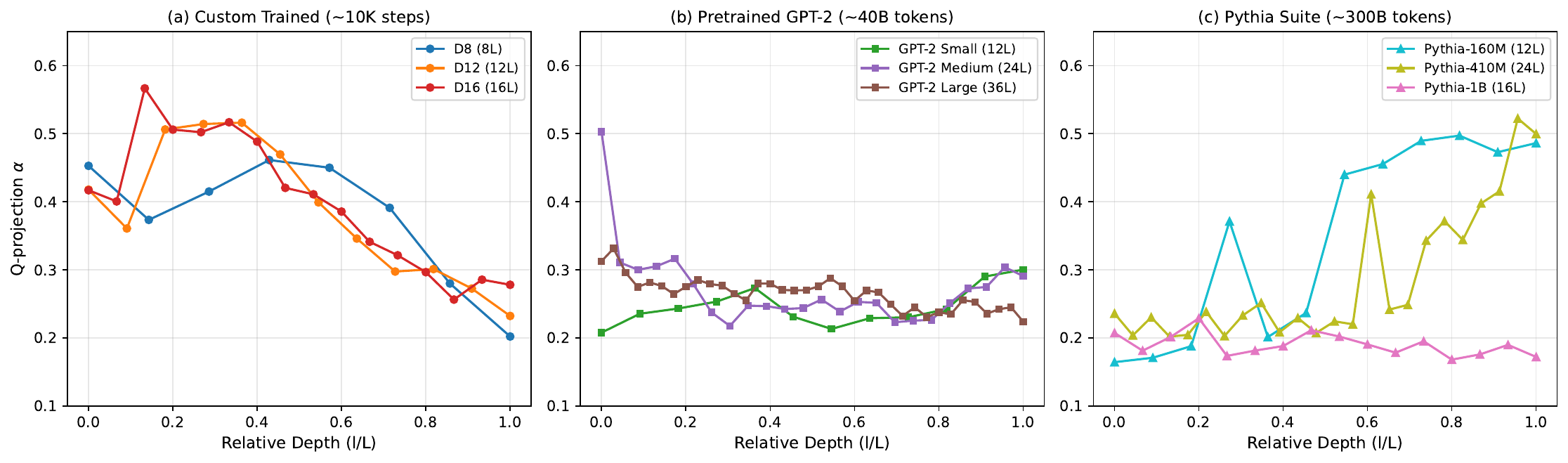}
    \caption{\textbf{Cross-family Q-$\alpha$ profiles (normalized depth).} (a)~Custom models: inverted-U with peak shifting left. (b)~GPT-2: early-layer concentration after extended training. (c)~Pythia: late-layer peaks at intermediate training. All show non-zero $\alpha$ gradients.}
    \label{fig:cross_family}
    \vspace{-5pt}
\end{figure}

Figure~\ref{fig:cross_family} summarizes nine models across three families. Three phenomena are universal: (1)~non-zero $\alpha$ gradients, (2)~Q/K--V asymmetry, and (3)~compression wave signatures. The $\alpha$-peak position depends strongly on training duration: short training $\to$ middle layers (our models), intermediate $\to$ late layers (Pythia), long training $\to$ early layers (GPT-2). Under controlled conditions, the power-law $\Delta\alpha \propto L^{0.26}$ holds tightly ($R^2{=}0.99$), but width and training duration modulate the relationship across families (Appendix~\ref{app:cross_family}).

\section{Practical Applications}
\label{sec:applications}

\subsection{Spectral $\alpha$ Predicts Layer Importance}
\label{sec:layer_importance}

We evaluate single-layer ablation (replacing each layer with identity) across all three model scales. Two patterns emerge:

\textbf{Boundary layers are irreplaceable:} L0--L1 cause catastrophic loss increases ($\Delta > 1.0$) regardless of $\alpha$, performing structurally constrained input processing.

\textbf{Among core layers, $\alpha$ predicts importance:} Spearman $\rho(\alpha, \Delta\text{Loss})$: 0.71 (D8), \textbf{0.84} ($p{=}0.002$, D12), 0.69 ($p{=}0.019$, D16 core L2--L12). High-$\alpha$ layers cause larger loss increases; low-$\alpha$ layers are nearly redundant ($\Delta < 0.01$). This reveals a \textbf{three-zone architecture}: input boundary / $\alpha$-predicted core / output boundary.

\subsection{Spectral-Guided Layer Pruning}
\label{sec:pruning}

We compare five strategies for removing $k \in \{1,2,3,4\}$ core layers: spectral-guided (lowest $\alpha$), Last-N~\citep{men2024shortgpt}, magnitude-based, random, and spectral-worst (highest $\alpha$, control). While \citet{lu2024alphapruning} used $\alpha$ to allocate \emph{intra-layer sparsity ratios}, we use it for \emph{inter-layer removal decisions}---a complementary application enabled by our discovery that $\alpha$ gradients correlate with functional importance.

Spectral-guided matches or outperforms all baselines on our custom models (Table~\ref{tab:pruning}). In D12 at $k{=}4$: spectral $\Delta{=}+0.220$ vs.\ spectral-worst $\Delta{=}+0.460$ ($2.1\times$ gap). However, on D16, spectral ordering coincides with Last-N because the lowest-$\alpha$ layers happen to be the last layers.

\textbf{Discriminative validation on GPT-2.} To confirm that $\alpha$ provides information \emph{beyond} layer position, we evaluate on GPT-2 Medium (24L, 355M) and GPT-2 Large (36L, 774M)---where spectral and Last-N orderings differ substantially (Table~\ref{tab:gpt2_pruning}). On GPT-2 Medium at $k{=}4$, spectral-guided achieves $\Delta\text{PPL}{=}+9.65$ vs.\ Last-N $\Delta{=}+25.06$ (\textbf{2.6$\times$ better}), while spectral-worst causes $\Delta{=}+229.15$ (\textbf{23.7$\times$ worse}). On GPT-2 Large, spectral outperforms Last-N by $1.5$--$2.0\times$ across all $k$ values, with worst-vs-best gaps reaching $20.4\times$ at $k{=}2$. This proves $\alpha$ captures genuine structural information beyond positional heuristics. The spectral ordering targets the $\alpha$ trough (depth 0.7--0.8 in Medium; L25--L30 in Large), while Last-N removes the true end layers including structurally important boundary layers. Full results including GPT-2 Small and extended $k{\in}\{6,8\}$ for Large are in Appendix~\ref{app:gpt2_pruning}.

\textbf{Cross-family validation on Pythia.} We extend pruning experiments to the full Pythia family (160M/410M/1B). Critically, Pythia models exhibit \emph{monotonically rising} $\alpha$ profiles (high-$\alpha$ layers are late)---the opposite topology from GPT-2's early peaks. This requires \textbf{zone-aware} spectral pruning (Algorithm~\ref{alg:zoneaware}) that protects boundary layers while targeting the interior low-$\alpha$ trough. On Pythia-1B (16L), zone-aware outperforms Last-N by $1.3$--$3.6\times$ ($k{=}2$--$4$), with worst-vs-best ratios exceeding $7\times$. Pythia-160M (12L) shows zone-aware outperforming Last-N by $1.1$--$2.0\times$ ($k{=}2$--$3$), while Pythia-410M (24L) reveals a crossover: Last-N wins at small $k$ but zone-aware dominates at aggressive pruning ($k{\geq}6$: $1.2$--$2.0\times$). This topology dependence confirms that $\alpha$ encodes \emph{structural}, not merely positional, information---the optimal pruning strategy depends on where the $\alpha$ gradient peaks, which varies across model families. Figure~\ref{fig:crossfamily_pruning} synthesizes results across all seven models. Full results in Appendix~\ref{app:pythia_pruning}.

\begin{table}[t]
\centering
\caption{Custom model pruning: $\Delta$Loss by strategy. Spectral$\,{\approx}\,$Last-N on D16 (coincident ordering).}
\label{tab:pruning}
\vspace{-3pt}
\small
\begin{tabular}{llccccc}
\toprule
Model & $k$ & Spectral & Last-N & Magnitude & Random & Worst \\
\midrule
\multirow{4}{*}{D12} 
& 1 & \textbf{+0.010} & +0.010 & +0.010 & +0.049 & +0.052 \\
& 2 & \textbf{+0.036} & +0.046 & +0.046 & +0.128 & +0.166 \\
& 3 & \textbf{+0.102} & +0.102 & +0.102 & +0.221 & +0.330 \\
& 4 & \textbf{+0.220} & +0.220 & +0.220 & +0.356 & +0.460 \\
\midrule
\multirow{4}{*}{D16}
& 1 & \textbf{+0.009} & +0.009 & +0.009 & +0.023 & +0.082 \\
& 2 & \textbf{+0.038} & +0.038 & +0.038 & +0.060 & +0.088 \\
& 3 & \textbf{+0.077} & +0.077 & +0.077 & +0.096 & +0.177 \\
& 4 & \textbf{+0.147} & +0.147 & +0.147 & +0.130 & +0.199 \\
\bottomrule
\end{tabular}
\vspace{-5pt}
\end{table}

\begin{table}[t]
\centering
\caption{GPT-2 pruning ($\Delta$PPL): spectral vs.\ Last-N orderings \textbf{differ}. Spectral-guided achieves 1.5--2.8$\times$ less degradation than Last-N across both model scales.}
\label{tab:gpt2_pruning}
\vspace{-3pt}
\small
\begin{tabular}{llccccc}
\toprule
Model & $k$ & Spectral & Last-N & Random & Worst & L-N/S \\
\midrule
\multirow{4}{*}{\shortstack{Medium\\(24L)}} 
& 1 & \textbf{+0.84} & +1.98 & +1.06 & +0.73 & 2.4$\times$ \\
& 2 & +2.67 & +6.04 & \textbf{+2.36} & +2.66 & 2.3$\times$ \\
& 3 & +4.89 & +13.49 & \textbf{+3.79} & +19.49 & 2.8$\times$ \\
& 4 & \textbf{+9.65} & +25.06 & +5.62 & +229.15 & 2.6$\times$ \\
\midrule
\multirow{4}{*}{\shortstack{Large\\(36L)}}
& 1 & \textbf{+0.50} & +0.75 & +0.44 & +0.62 & 1.5$\times$ \\
& 2 & \textbf{+1.04} & +1.92 & +1.06 & +21.20 & 1.9$\times$ \\
& 3 & \textbf{+1.68} & +3.41 & +1.51 & +26.80 & 2.0$\times$ \\
& 4 & \textbf{+2.96} & +5.55 & +2.18 & +33.36 & 1.9$\times$ \\
\bottomrule
\end{tabular}
\vspace{-5pt}
\end{table}

\subsection{Spectral Warmup: A Revealing Negative Result}
\label{sec:warmup}

We test \emph{Spectral Warmup}: initializing weight spectra to their post-training targets using random orthogonal directions. Despite starting with the ``correct'' spectral distribution, this trains \textbf{42.7\% worse} than standard initialization. This demonstrates that singular value \emph{directions} ($\mathbf{U}$, $\mathbf{V}$) encode the vast majority of learned information. Correct spectral shape with random directions is worse than random initialization---directions must co-evolve with magnitudes. Details in Appendix~\ref{app:warmup}.

\begin{algorithm}[t]
\caption{Zone-Aware Spectral Layer Pruning}
\label{alg:zoneaware}
\begin{algorithmic}[1]
\REQUIRE Model $\mathcal{M}$ with $L$ layers, target removal count $k$, boundary size $b$
\STATE \textbf{Compute} $\alpha_l$ for each layer $l \in \{0, \ldots, L{-}1\}$ via weighted power-law fit
\STATE \textbf{Define} interior $\mathcal{I} = \{l : b \leq l < L{-}b\}$, boundary $\mathcal{B} = \{0,\ldots,b{-}1\} \cup \{L{-}b,\ldots,L{-}1\}$
\STATE \textbf{Sort} interior layers by $\alpha$: $\mathcal{I}_{\text{sorted}} = \text{argsort}_{l \in \mathcal{I}}(\alpha_l)$
\STATE \textbf{Select} $k$ layers from $\mathcal{I}_{\text{sorted}}$ with minimum gap $\geq 2$
\STATE \textbf{Remove} selected layers (replace with identity)
\RETURN Pruned model
\end{algorithmic}
\end{algorithm}

\begin{figure}[t]
    \centering
    \includegraphics[width=\linewidth]{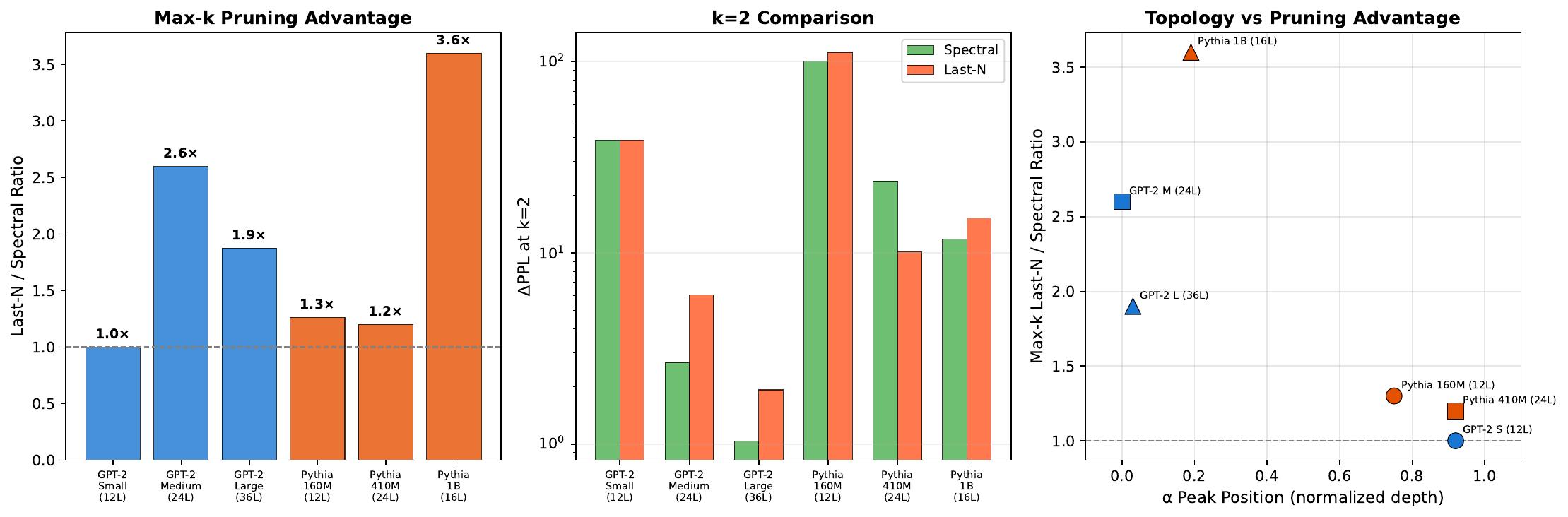}
    \caption{\textbf{Cross-family pruning synthesis (7 models, 2 families).} (a)~Last-N/Spectral ratio at maximum $k$: spectral wins on 5/7 models. (b)~$\Delta$PPL at $k{=}2$ across all models. (c)~Topology dependence: the $\alpha$-peak position modulates pruning advantage. Models with mid-network peaks (GPT-2 Medium, Pythia-1B) show the largest spectral advantage.}
    \label{fig:crossfamily_pruning}
    \vspace{-5pt}
\end{figure}

\section{Conclusion}
\label{sec:conclusion}

We have presented the first spatiotemporal study of weight matrix spectral dynamics during transformer pretraining, analyzing over 150{,}000 SVD snapshots across nine models in three families. Where prior work~\citep{martin2021implicit, yunis2024spectral} analyzed individual matrices or static snapshots, our inter-layer temporal analysis reveals that rank and spectral shape encode fundamentally different information---a dissociation invisible without the spatiotemporal perspective. The two-timescale theory, validated by scaling laws ($\Delta\alpha \propto L^{0.26}$) and cross-family confirmation, establishes that fast compression (how much structure) and slow shape differentiation (what kind) are the two fundamental axes of spectral learning in transformers.

The practical implications---spectral-guided pruning ($1.1$--$3.6\times$ better than Last-N across seven models in two families: GPT-2 124M--774M and Pythia 160M--1B), layer importance prediction, and the spectral warmup negative result---demonstrate that these theoretical insights translate into actionable tools. The three-zone architecture (input boundary / $\alpha$-predicted core / output boundary) provides a principled framework for where spectral metrics are and are not informative.

The finding that Q/K projections carry depth-dependent dynamics while V/O are uniform suggests a fundamental architectural principle: \emph{attention selection is the adaptive computation}, while value transformation is generic. This aligns with attention head specialization (induction heads, positional heads)~\citep{olsson2022context}. The persistent $\alpha$ gradient provides a \emph{quantitative} signature of functional specialization from weights alone, without requiring input data---complementing probing~\citep{tenney2019bert} and activation analysis~\citep{geva2022transformer}.

\paragraph{Limitations.} Full SVD tracking at multi-billion scale requires efficient approximations~\citep{halko2011finding}. The D16 model reached 8{,}970 of 10{,}000 target steps. Layer importance for D8 ($p{=}0.11$) does not reach significance due to only 6 interior layers. The custom tokenizer (8K vocab) may influence early dynamics.

\paragraph{Future Work.} Spectral-aware per-layer learning rates; extension to MoE, SSMs, and vision architectures; Spectral Warmup with direction transfer; real-time spectral diagnostics for production training.

\paragraph{Broader Impact.} This work is foundational research on understanding transformer training dynamics. The primary societal benefit is enabling more efficient model training and compression, reducing computational costs and energy consumption. We do not foresee direct negative societal impacts. Our spectral monitoring tools are diagnostic and do not introduce new capabilities for harmful applications.

\paragraph{Reproducibility.} All custom model training uses publicly available data (ClimbMix-400B) with complete hyperparameters in Appendix~\ref{app:details}. Pretrained model analysis uses publicly available checkpoints (GPT-2 from HuggingFace, Pythia from EleutherAI). All spectral metrics are computed via standard SVD (\texttt{torch.linalg.svdvals}). Multi-seed experiments ($n{=}4$) confirm all qualitative findings. Code for spectral analysis, training, and figure generation will be released upon publication.

\bibliographystyle{plainnat}
\bibliography{references}

\newpage
\appendix

\section{Extended Experimental Details}
\label{app:details}

\subsection{Data Pipeline}
We use ClimbMix-400B~\citep{karpathy2024}, a shuffled web text corpus. We train a custom BPE tokenizer with vocabulary size 8{,}192 on a 100M character subset. Documents are packed into fixed-length sequences of 2{,}048 tokens using a best-fit algorithm. Each sequence begins with a BOS token.

\subsection{Training Hyperparameters}

\begin{table}[ht]
\centering
\caption{Complete training hyperparameters.}
\label{tab:hyperparams}
\small
\begin{tabular}{lc}
\toprule
Hyperparameter & Value \\
\midrule
Optimizer & AdamW \\
$\beta_1, \beta_2$ & 0.9, 0.95 \\
Weight decay & 0.1 \\
Gradient clipping & 1.0 \\
Learning rate & $6 \times 10^{-4}$ \\
Schedule & Cosine with 200-step warmup \\
Batch size & 8 (D8/D12), 4 (D16) \\
Sequence length & 2{,}048 \\
Tokens per step & 16{,}384 (D8/D12), 8{,}192 (D16) \\
Total training tokens & ${\sim}$164M (10K steps) \\
SVD interval & 25 steps (D8/D12), 50 steps (D16) \\
Vocabulary size & 8{,}192 \\
Activation/Norm & GELU / Pre-LayerNorm \\
Bias / Weight tying & None / Yes \\
\bottomrule
\end{tabular}
\end{table}

\subsection{SVD Computation}
Full SVDs computed via \texttt{torch.linalg.svdvals} in float32. Total across all experiments: $>$150{,}000 SVD snapshots. Monitoring overhead: $<$5\% of training time.

\section{Spectral Monitoring Protocol}
\label{app:protocol}

\begin{algorithm}[t]
\caption{Spectral Monitoring During Training}
\label{alg:spectral_monitoring}
\begin{algorithmic}[1]
\REQUIRE Model $\mathcal{M}$ with layers $\{l_0, \ldots, l_{L-1}\}$, interval $\Delta t$, types $\mathcal{T} = \{Q, K, V, O, \text{MLP}_\uparrow, \text{MLP}_\downarrow\}$
\STATE Initialize spectral log $\mathcal{S} \leftarrow \emptyset$
\FOR{each training step $t = 0, \Delta t, 2\Delta t, \ldots$}
    \FOR{each layer $l$, each type $\tau \in \mathcal{T}$}
        \STATE $\boldsymbol{\sigma} \leftarrow \textsc{SvdVals}(\mathcal{M}[l, \tau])$
        \STATE Compute $R_s$, $\alpha$, $H$, spectral gap
        \STATE $\mathcal{S} \leftarrow \mathcal{S} \cup \{(t, l, \tau, R_s, \alpha, H, \sigma_1/\sigma_2)\}$
    \ENDFOR
\ENDFOR
\RETURN $\mathcal{S}$
\end{algorithmic}
\end{algorithm}

\section{Q/K--V Functional Asymmetry: Full Results}
\label{app:qkv}

Table~\ref{tab:qkv_full} presents the complete Q/K--V comparison across all model scales.

\begin{table}[ht]
\centering
\caption{Q/K--V asymmetry. V/O projections compress uniformly ($>$90\%), while Q/K exhibit depth-dependent dynamics.}
\label{tab:qkv_full}
\small
\begin{tabular}{lcccccc}
\toprule
& \multicolumn{2}{c}{Mean Compression} & \multicolumn{2}{c}{$\alpha$ Range} & \multicolumn{2}{c}{$\alpha$ Std} \\
\cmidrule(lr){2-3} \cmidrule(lr){4-5} \cmidrule(lr){6-7}
Model & Q & V & Q & V & Q & V \\
\midrule
D8  & 91.2\% & 95.1\% & 0.26 & 0.08 & 0.095 & 0.025 \\
D12 & 89.7\% & 94.8\% & 0.28 & 0.10 & 0.098 & 0.031 \\
D16 & 85.4\% & 93.2\% & 0.31 & 0.09 & 0.102 & 0.028 \\
\bottomrule
\end{tabular}
\end{table}

\begin{figure}[ht]
    \centering
    \includegraphics[width=\linewidth]{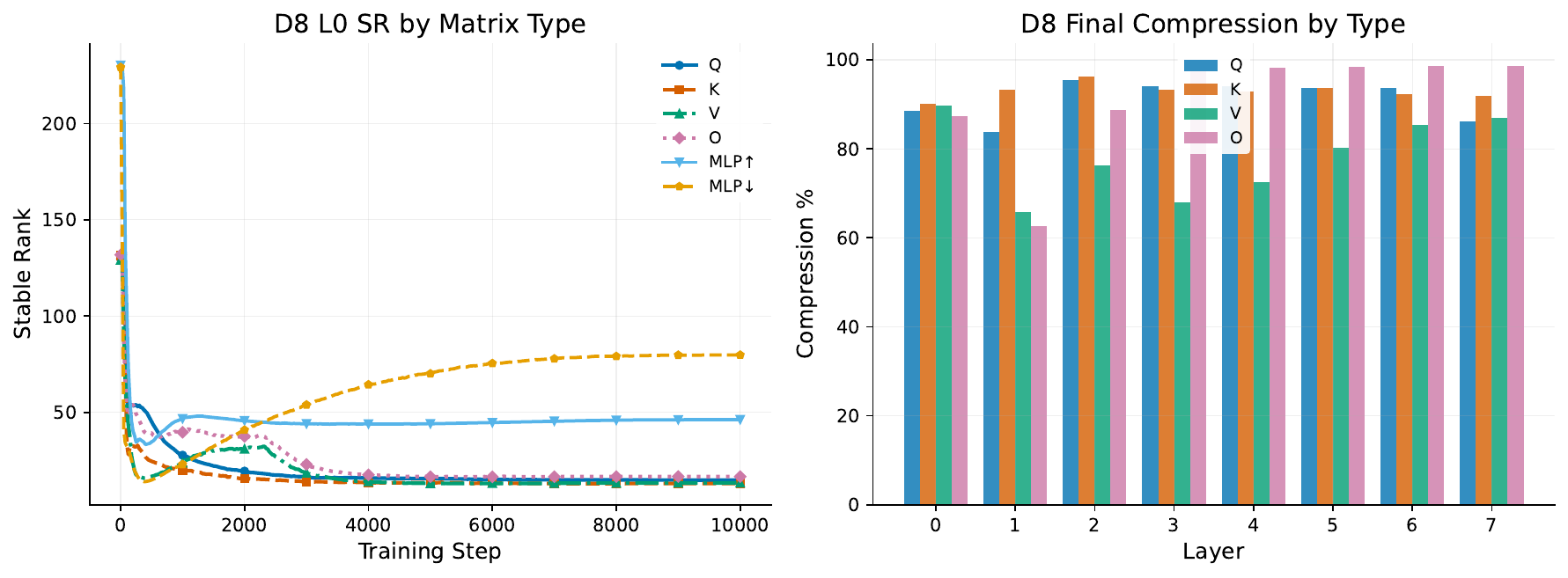}
    \caption{Q/K vs.\ V $\alpha$ profiles. Q/K projections carry depth-dependent dynamics; V projections are nearly flat.}
    \label{fig:qkv_asymmetry_app}
\end{figure}

Figure~\ref{fig:qkv_asymmetry_app} shows the per-matrix-type comparison.

\section{Multi-Seed Reproducibility}
\label{app:reproducibility}

We validate all findings with $n{=}4$ random seeds for both D8 and D16.

\begin{figure}[ht]
    \centering
    \begin{subfigure}[t]{0.48\linewidth}
        \includegraphics[width=\linewidth]{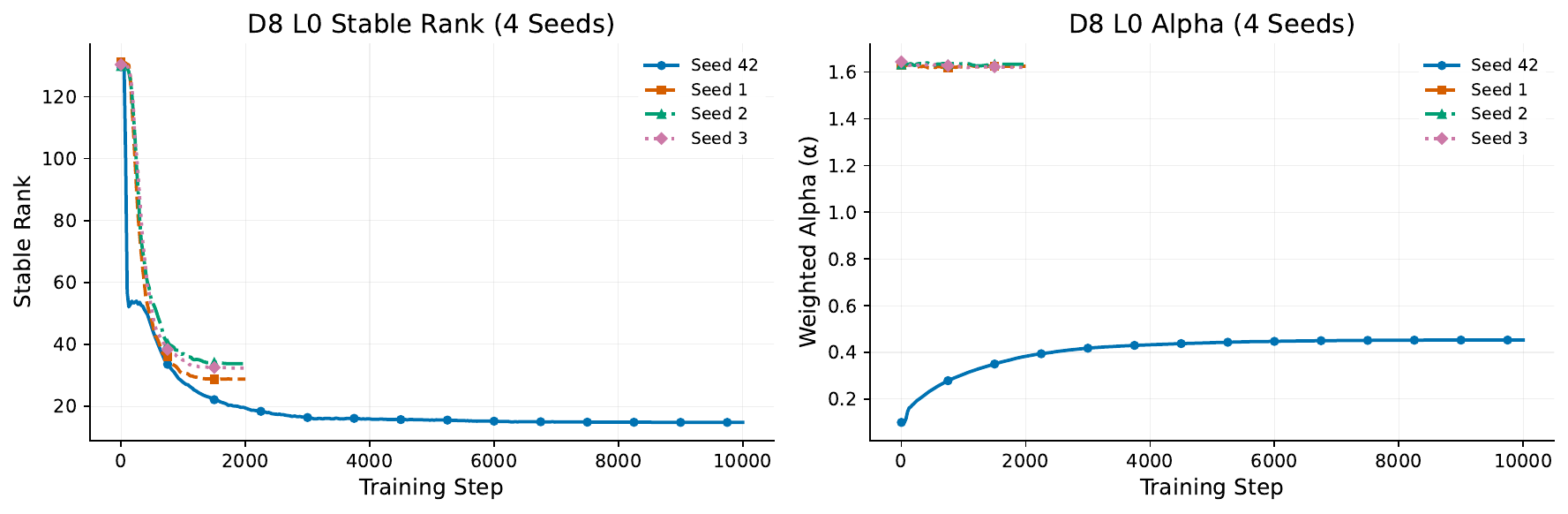}
        \caption{D8 multi-seed ($n{=}4$).}
    \end{subfigure}
    \hfill
    \begin{subfigure}[t]{0.48\linewidth}
        \includegraphics[width=\linewidth]{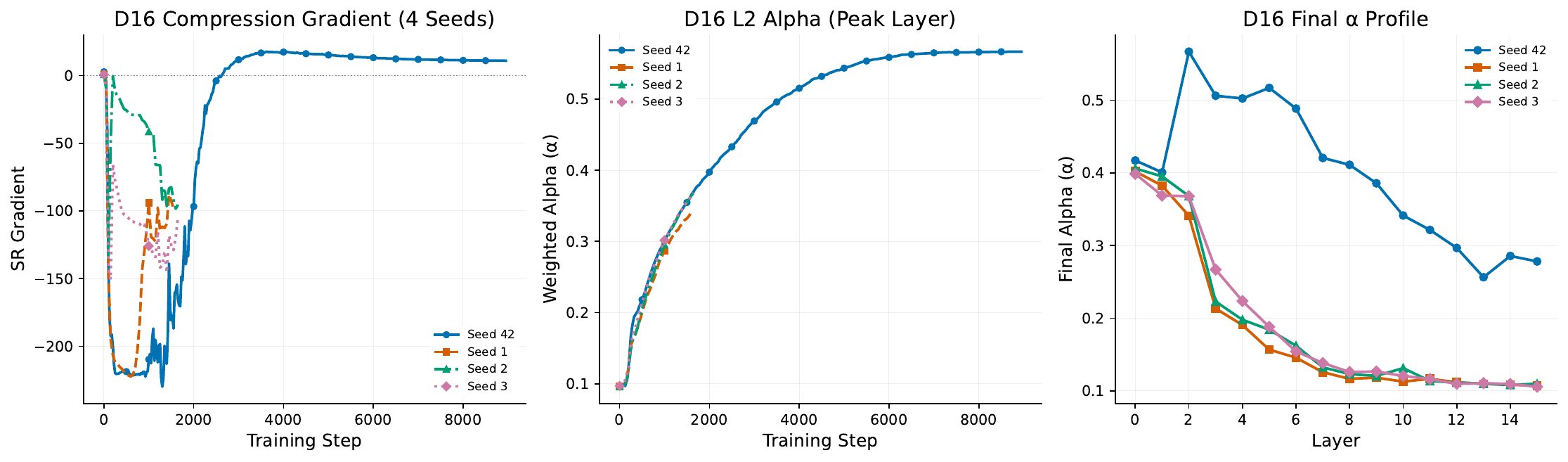}
        \caption{D16 multi-seed ($n{=}4$).}
    \end{subfigure}
    \caption{Multi-seed reproducibility. All qualitative phenomena (compression waves, $\alpha$ gradients, inverted-U) are consistent across seeds.}
    \label{fig:multiseed_app}
\end{figure}

\section{Spectral Entropy Confirmation}
\label{app:entropy}

\begin{figure}[ht]
    \centering
    \includegraphics[width=\linewidth]{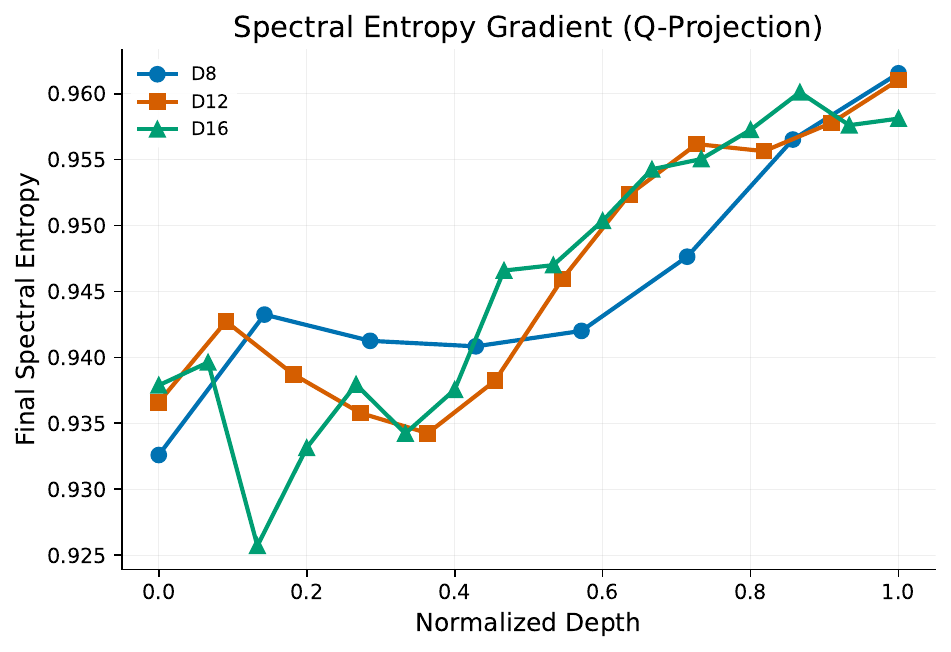}
    \caption{Spectral entropy gradient mirrors $\alpha$: lower entropy (more concentrated spectra) at early-middle layers. This independently confirms the persistent gradient.}
    \label{fig:entropy_app}
\end{figure}

\section{Spectral Warmup: Full Details}
\label{app:warmup}

Spectral Warmup initializes each weight matrix as $\mathbf{W}_0 = \mathbf{U}_{\text{rand}} \cdot \text{diag}(s \cdot \boldsymbol{\sigma}^*) \cdot \mathbf{V}_{\text{rand}}^\top$ where $\boldsymbol{\sigma}^*$ are target singular values from a trained reference model and $\mathbf{U}_{\text{rand}}, \mathbf{V}_{\text{rand}}$ are random orthogonal matrices.

\begin{table}[ht]
\centering
\caption{Spectral Warmup vs.\ Standard Init (D8, 5K steps).}
\label{tab:warmup_app}
\small
\begin{tabular}{lcccc}
\toprule
Method & Val Loss @5K & Initial SR & Initial $\alpha$ & Final SR \\
\midrule
Standard & \textbf{3.720} & ${\sim}$130 & ${\sim}$0.10 & ${\sim}$15 \\
Warmup & 5.307 & ${\sim}$15 & ${\sim}$0.45 & ${\sim}$22 \\
\bottomrule
\end{tabular}
\end{table}

The 42.7\% gap demonstrates that directions $\gg$ magnitudes: correct spectral shape with random directions is far worse than random init with incorrect spectra.

\section{Layer Importance: Full Results}
\label{app:importance}

\begin{figure}[ht]
    \centering
    \includegraphics[width=\linewidth]{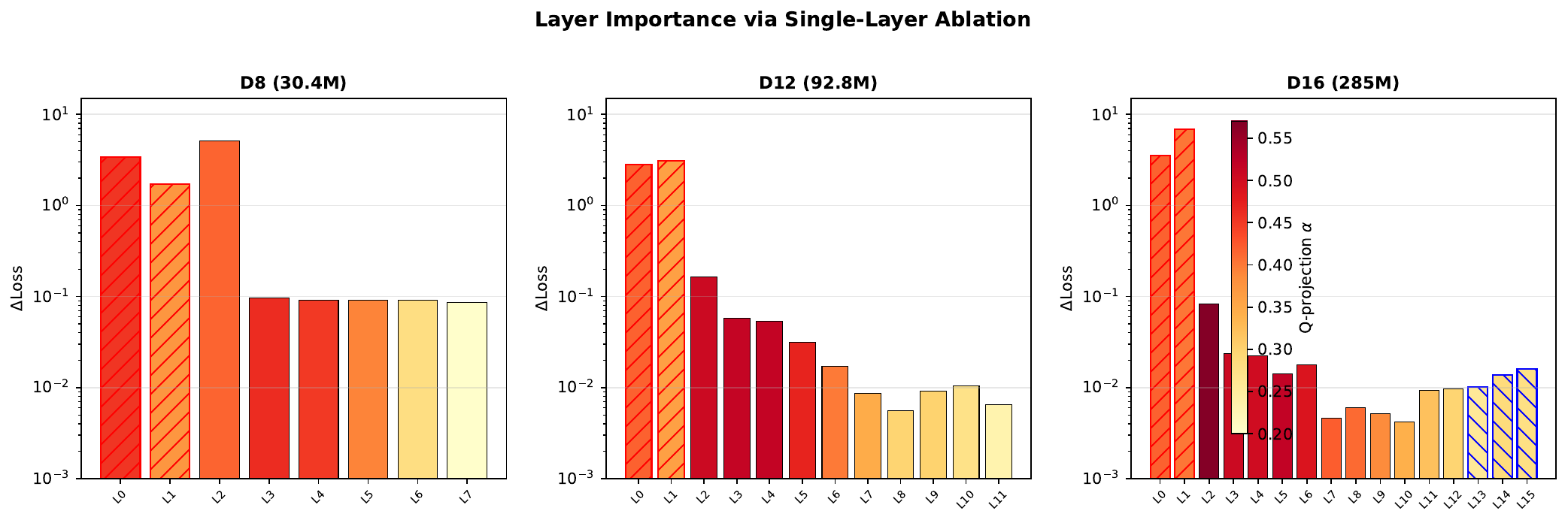}
    \caption{Layer importance via ablation. Color = $\alpha$. Among core layers, higher $\alpha$ $\to$ higher importance.}
    \label{fig:importance_app}
\end{figure}

\begin{figure}[ht]
    \centering
    \includegraphics[width=\linewidth]{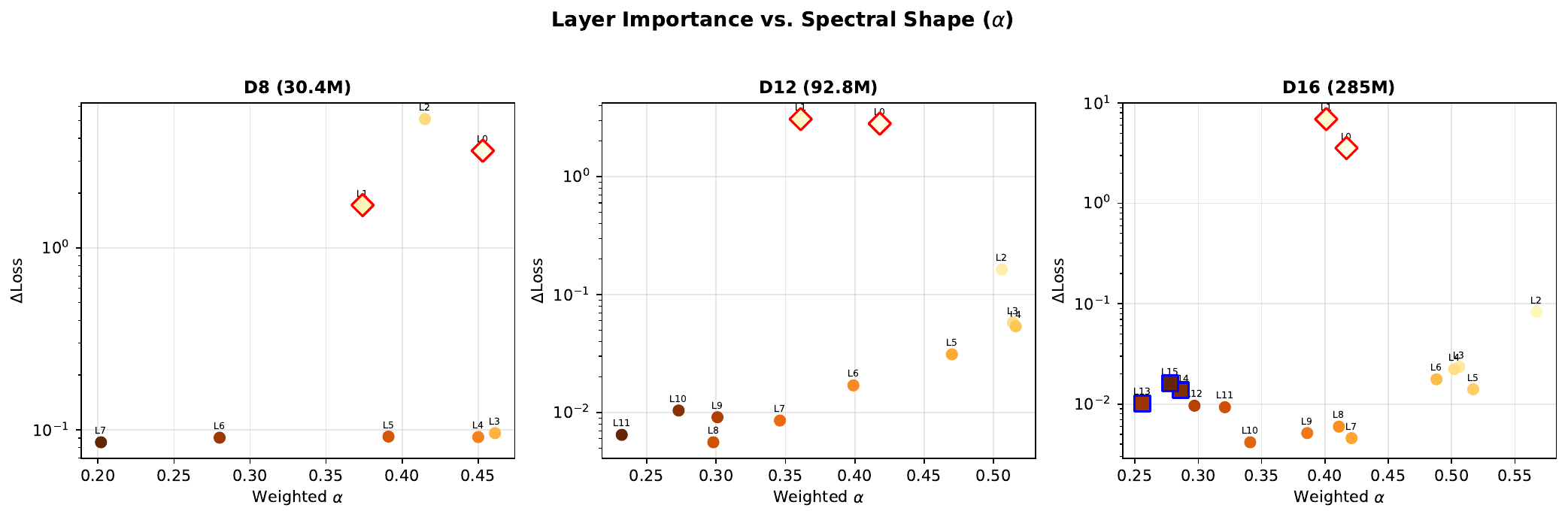}
    \caption{Layer importance--$\alpha$ scatter plots. Boundary layers are structural outliers; core layers show positive $\alpha$--importance correlation.}
    \label{fig:importance_scatter_app}
\end{figure}

\begin{table}[ht]
\centering
\caption{Full layer ablation results.}
\label{tab:importance_app}
\small
\begin{tabular}{lcccccc}
\toprule
& \multicolumn{2}{c}{D8} & \multicolumn{2}{c}{D12} & \multicolumn{2}{c}{D16} \\
\cmidrule(lr){2-3} \cmidrule(lr){4-5} \cmidrule(lr){6-7}
Layer & $\alpha$ & $\Delta$Loss & $\alpha$ & $\Delta$Loss & $\alpha$ & $\Delta$Loss \\
\midrule
\rowcolor{gray!15} L0 & 0.453 & +3.42 & 0.418 & +2.82 & 0.417 & +3.56 \\
\rowcolor{gray!15} L1 & 0.374 & +1.72 & 0.361 & +3.08 & 0.401 & +6.91 \\
L2  & 0.415 & +5.10 & 0.506 & +0.163 & 0.567 & +0.083 \\
L3  & 0.461 & +0.096 & 0.514 & +0.058 & 0.506 & +0.024 \\
L4  & 0.450 & +0.091 & 0.516 & +0.054 & 0.502 & +0.022 \\
L5  & 0.391 & +0.092 & 0.470 & +0.031 & 0.517 & +0.014 \\
L6  & 0.280 & +0.090 & 0.399 & +0.017 & 0.488 & +0.018 \\
L7  & 0.202 & +0.085 & 0.346 & +0.009 & 0.421 & +0.005 \\
L8  &  &  & 0.298 & +0.006 & 0.411 & +0.006 \\
L9  &  &  & 0.301 & +0.009 & 0.386 & +0.005 \\
L10 &  &  & 0.273 & +0.010 & 0.341 & +0.004 \\
L11 &  &  & 0.232 & +0.007 & 0.321 & +0.009 \\
L12 &  &  &  &  & 0.297 & +0.010 \\
\rowcolor{gray!15} L13 &  &  &  &  & 0.256 & +0.010 \\
\rowcolor{gray!15} L14 &  &  &  &  & 0.286 & +0.014 \\
\rowcolor{gray!15} L15 &  &  &  &  & 0.278 & +0.016 \\
\midrule
$\rho$ [L2+] & \multicolumn{2}{c}{0.71} & \multicolumn{2}{c}{0.84$^{**}$} & \multicolumn{2}{c}{0.44} \\
$\rho$ [core] &  &  &  &  & \multicolumn{2}{c}{0.69$^{*}$} \\
\bottomrule
\multicolumn{7}{l}{\small $^{**}p{<}0.01$; $^{*}p{<}0.05$; core=L2--L12.}
\end{tabular}
\end{table}

\section{Cross-Family Synthesis}
\label{app:cross_family}

This appendix consolidates the full cross-family comparison that is summarized in Figure~\ref{fig:cross_family}. The goal is to make explicit how the three model families differ not only in absolute $\Delta\alpha$ magnitude, but also in \emph{where} along depth the spectral peak appears and how that peak shifts with training duration.

Three patterns are worth highlighting. First, the custom short-training models (D8/D12/D16) show relatively large $\Delta\alpha$ values together with peaks in the early-to-middle layers, consistent with the main-text claim that partially trained models preserve a strong interior specialization gradient. Second, the GPT-2 family exhibits much smaller or less stable cross-layer separation in the final checkpoint, and its peak location shifts dramatically toward the earliest layers in medium and large models, matching the interpretation that long training drives the system toward early-layer spectral concentration. Third, the Pythia family occupies a distinct late-peaking regime: Pythia-160M and Pythia-410M both reach their strongest spectral separation in the final quarter of the network, whereas Pythia-1B flattens substantially despite similar training duration. This makes Pythia-1B an informative exception rather than a contradiction, suggesting that width and optimization trajectory can partially erase the depth gradient even when the family-level tendency remains late-peaking.

Taken together, the table below is the appendix-level evidence behind the main-text claim that \emph{training duration determines peak direction, while architecture and scale modulate peak sharpness}. In other words, the cross-family story is not a single universal curve, but a structured phase diagram: short-trained custom models peak in the interior, very long-trained GPT-2 models peak early, and intermediate-duration Pythia checkpoints peak late.

\begin{table}[ht]
\centering
\caption{Complete cross-family comparison: nine models, three families, 30M--1B parameters.}
\label{tab:cross_family_app}
\small
\begin{tabular}{llcccccc}
\toprule
Family & Model & $L$ & Params & $\Delta\alpha$ & Peak $l^*$ & $l^*/L$ & Training \\
\midrule
Custom & D8 & 8 & 30M & 0.259 & L3 & 0.38 & 10K steps \\
& D12 & 12 & 93M & 0.284 & L4 & 0.33 & 10K steps \\
& D16 & 16 & 285M & 0.310 & L2 & 0.13 & 10K steps \\
\midrule
GPT-2 & Small & 12 & 124M & 0.092 & L11 & 0.92 & ${\sim}$40B tok \\
& Medium & 24 & 355M & 0.285 & L0 & 0.00 & ${\sim}$40B tok \\
& Large & 36 & 774M & 0.107 & L1 & 0.03 & ${\sim}$40B tok \\
\midrule
Pythia & 160M & 12 & 160M & 0.333 & L9 & 0.75 & 143K steps \\
& 410M & 24 & 410M & 0.320 & L22 & 0.92 & 143K steps \\
& 1B & 16 & 1B & 0.061 & L3 & 0.19 & 143K steps \\
\bottomrule
\end{tabular}
\end{table}

A final practical takeaway is that transfer across families should be done with care. A pruning, monitoring, or interpretability heuristic calibrated on GPT-2 is likely to fail if applied unchanged to Pythia, because the low-$\alpha$ and high-$\alpha$ regions occupy different functional zones of the depth axis. This is precisely why the topology-aware pruning rules introduced later in the appendix are necessary: the spectral signal is meaningful across families, but its operational interpretation must respect family-specific geometry.

\section{GPT-2 Pruning: Full Results}
\label{app:gpt2_pruning}

Table~\ref{tab:gpt2_pruning_full} presents complete pruning results for GPT-2 Small, Medium, and Large. On GPT-2 Small (12L), spectral and Last-N orderings partially overlap (lowest-$\alpha$ layers L8--L10 are near the end), so discrimination is weaker at $k{\geq}2$. On GPT-2 Medium (24L), the orderings diverge substantially: spectral targets L17--L20 (the $\alpha$ trough at normalized depth 0.7--0.8), while Last-N targets L19--L22 (the true end). On GPT-2 Large (36L, 774M), with the most layers, the $\alpha$ trough is concentrated in L25--L30 while Last-N removes L27--L34. Spectral consistently outperforms Last-N by $1.5$--$2.0\times$, and the worst-vs-best gap reaches $20.4\times$ at $k{=}2$. For $k \in \{6, 8\}$---aggressive pruning of 17--22\% of layers---spectral still maintains its advantage, with worst-case PPL at $k{=}8$ reaching $+243.94$ ($14.8\times$ the spectral-best impact).

\begin{table}[ht]
\centering
\caption{Complete GPT-2 pruning results ($\Delta$PPL). Baseline PPL: Small = 24.36, Medium = 18.00, Large = 15.55.}
\label{tab:gpt2_pruning_full}
\small
\begin{tabular}{llccccc}
\toprule
Model & $k$ & Spectral & Last-N & Random & Worst & W/S Ratio \\
\midrule
\multirow{4}{*}{Small (12L)} 
& 1 & \textbf{+4.48} & +9.33 & +4.62 & +1.37 & 0.3$\times$ \\
& 2 & +38.72 & +38.72 & \textbf{+14.43} & +10.18 & 0.3$\times$ \\
& 3 & +93.72 & +93.72 & \textbf{+34.69} & +106.42 & 1.1$\times$ \\
& 4 & +259.53 & +259.53 & \textbf{+97.34} & +3318.96 & 12.8$\times$ \\
\midrule
\multirow{4}{*}{Medium (24L)}
& 1 & \textbf{+0.84} & +1.98 & +1.06 & +0.73 & 0.9$\times$ \\
& 2 & +2.67 & +6.04 & \textbf{+2.36} & +2.66 & 1.0$\times$ \\
& 3 & +4.89 & +13.49 & \textbf{+3.79} & +19.49 & 4.0$\times$ \\
& 4 & \textbf{+9.65} & +25.06 & +5.62 & +229.15 & 23.7$\times$ \\
\midrule
\multirow{6}{*}{Large (36L)}
& 1 & \textbf{+0.50} & +0.75 & +0.44 & +0.62 & 1.3$\times$ \\
& 2 & \textbf{+1.04} & +1.92 & +1.06 & +21.20 & 20.4$\times$ \\
& 3 & \textbf{+1.68} & +3.41 & +1.51 & +26.80 & 15.9$\times$ \\
& 4 & \textbf{+2.96} & +5.55 & +2.18 & +33.36 & 11.3$\times$ \\
& 6 & \textbf{+7.22} & +10.29 & +3.92 & +64.49 & 8.9$\times$ \\
& 8 & \textbf{+16.43} & +19.93 & +6.55 & +243.94 & 14.8$\times$ \\
\bottomrule
\end{tabular}
\end{table}

\begin{figure}[ht]
    \centering
    \includegraphics[width=\linewidth]{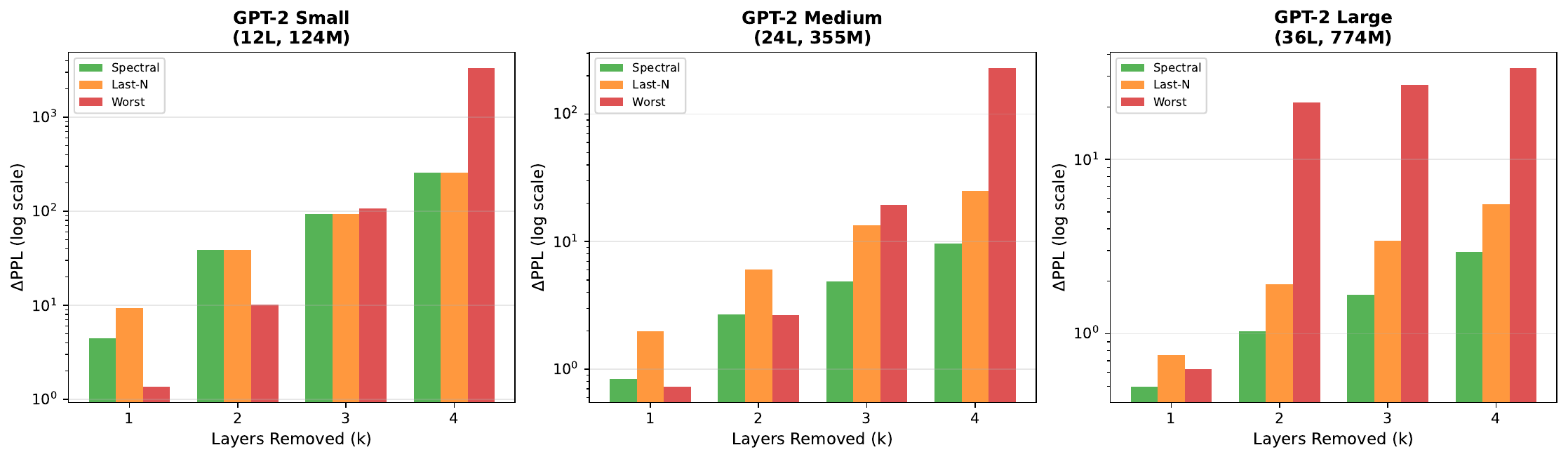}
    \caption{GPT-2 family pruning comparison. Left: Small (12L)---spectral and Last-N overlap at $k{\geq}2$. Center: Medium (24L)---spectral outperforms Last-N by $2.3$--$2.8\times$. Right: Large (36L)---spectral outperforms Last-N by $1.5$--$2.0\times$ with worst-vs-best gap up to $20.4\times$.}
    \label{fig:gpt2_pruning_app}
\end{figure}

\begin{figure}[ht]
    \centering
    \includegraphics[width=\linewidth]{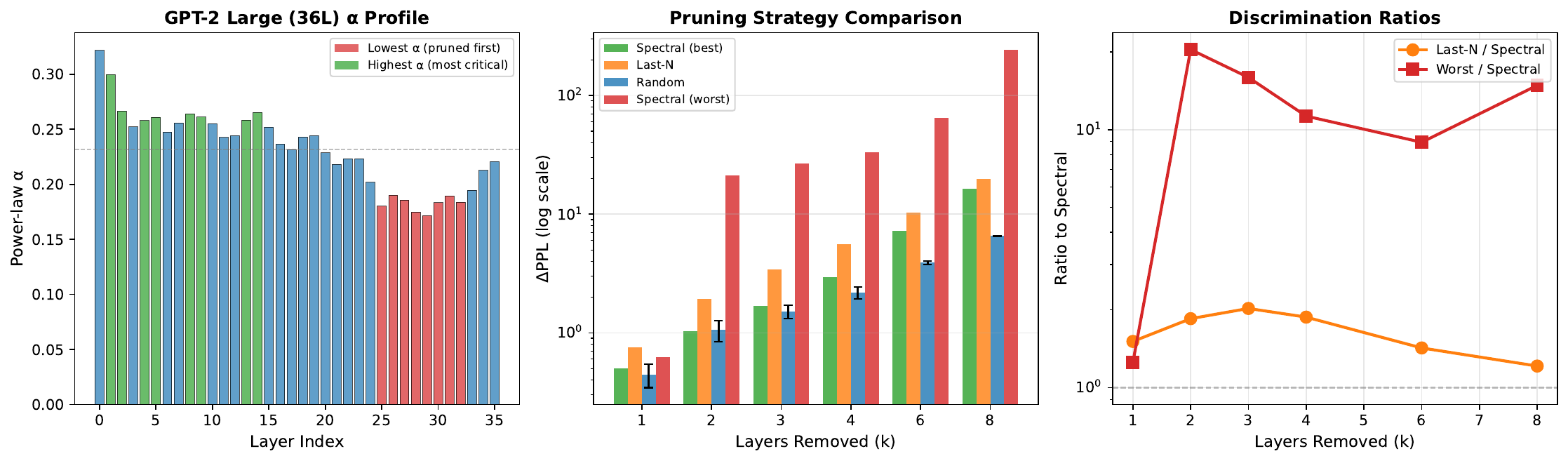}
    \caption{GPT-2 Large (36L, 774M) spectral pruning analysis. Left: $\alpha$ profile showing low-$\alpha$ layers (red, pruned first) concentrated in L25--L30, distinct from the Last-N targets. Center: pruning impact comparison on log scale. Right: discrimination ratios---Last-N/Spectral ratio stays $1.2$--$2.0\times$; Worst/Spectral reaches $20.4\times$ at $k{=}2$.}
    \label{fig:gpt2large_pruning_app}
\end{figure}

Note that on GPT-2 Small at $k{=}1$, spectral-worst (removing L1, $\alpha{=}0.237$) causes \emph{less} damage than spectral-best (removing L9, $\alpha{=}0.135$). This is because L1 is adjacent to the boundary zone where $\alpha$ is high due to proximity to the embedding, not functional importance---consistent with our three-zone architecture (Section~\ref{sec:layer_importance}).

\section{Pythia Pruning: Cross-Family Validation}
\label{app:pythia_pruning}

\begin{figure}[ht]
    \centering
    \includegraphics[width=\linewidth]{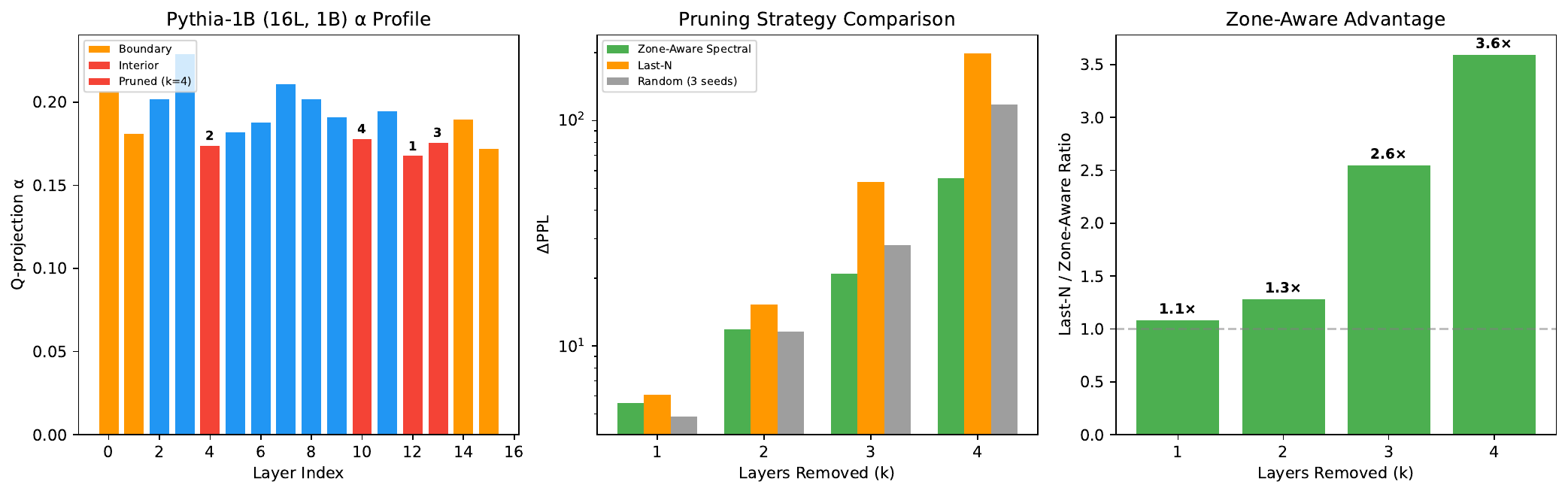}
    \caption{Pythia-1B (16L, 1B) zone-aware spectral pruning. (a)~$\alpha$ profile with boundary (orange) and interior layers; red bars indicate pruning targets (k=4). (b)~$\Delta$PPL comparison on log scale. (c)~Last-N/Zone-Aware ratio: advantage grows from $1.1\times$ to $3.6\times$ with increasing $k$.}
    \label{fig:pythia1b_pruning_app}
\end{figure}

The Pythia models exhibit \emph{monotonically rising} $\alpha$ profiles (Figure~\ref{fig:pythia1b_pruning_app}a)---the opposite topology from GPT-2's early-peak pattern. This creates a critical challenge for naive spectral pruning: the lowest-$\alpha$ layers are early foundational layers whose removal is catastrophic. We introduce \textbf{zone-aware} spectral pruning (Algorithm~\ref{alg:zoneaware}) that protects boundary layers and targets the interior low-$\alpha$ trough.

\subsection{Pythia-160M (12L, 160M Parameters)}

Pythia-160M has a rising $\alpha$ profile ($\alpha \in [0.165, 0.497]$, $\Delta\alpha{=}0.333$) with a steep transition at L5--L6. Zone-aware pruning (boundary$\,{=}\,$1) targets the low-$\alpha$ interior layers L1, L4, L5 while protecting L0 and L11.

At $k{=}3$, zone-aware achieves $\Delta\text{PPL}{=}+195.23$ vs.\ Last-N $\Delta{=}+389.77$ (\textbf{2.0$\times$ better}), while spectral-worst causes $\Delta{=}+156{,}415$ (\textbf{800$\times$ worse}). At $k{=}4$, the worst-vs-best ratio reaches an extraordinary $11{,}880\times$, demonstrating that even in 12-layer models, $\alpha$ captures critical structural information.

\begin{figure}[ht]
    \centering
    \includegraphics[width=\linewidth]{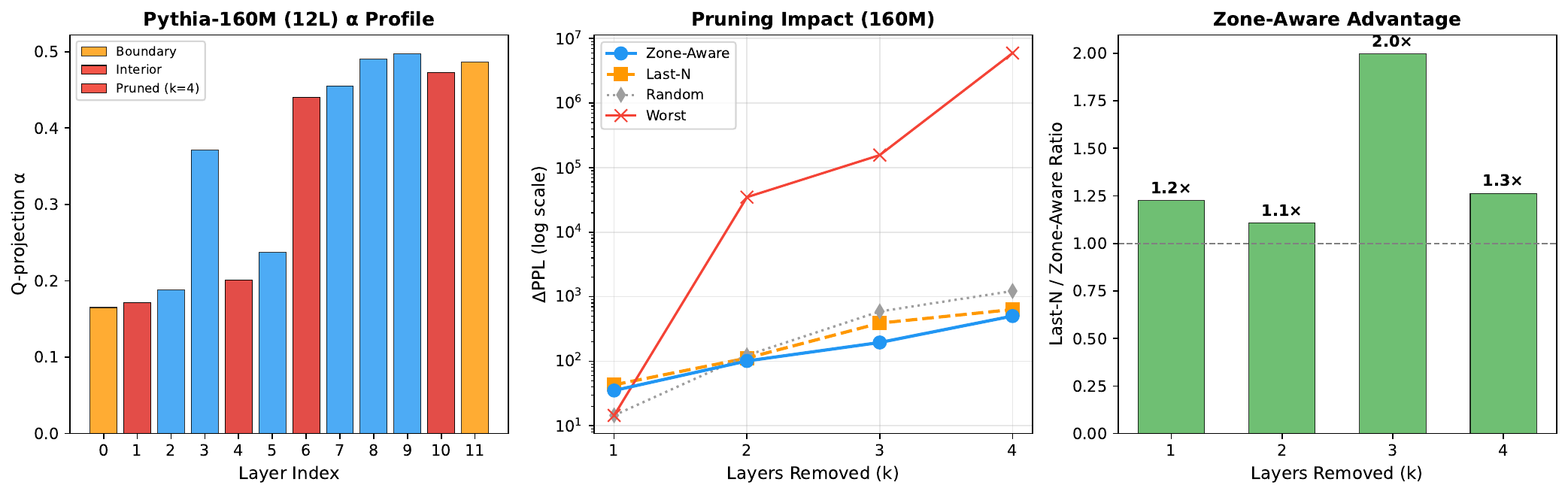}
    \caption{Pythia-160M (12L) zone-aware spectral pruning. (a)~$\alpha$ profile showing rising trend with pruning targets (red, k=4). (b)~$\Delta$PPL comparison on log scale. (c)~Zone-aware advantage grows with $k$.}
    \label{fig:pythia160m_pruning_app}
\end{figure}

\begin{table}[ht]
\centering
\caption{Pythia-160M pruning ($\Delta$PPL). Baseline PPL = 25.49.}
\label{tab:pythia160m_pruning}
\small
\begin{tabular}{lccccc}
\toprule
$k$ & Zone-Aware & Last-N & Random & Worst & L-N/ZA \\
\midrule
1 & +35.21 & +43.14 & +14.50 & +14.47 & 1.2$\times$ \\
2 & \textbf{+101.03} & +111.94 & +123.59 & +34{,}875 & 1.1$\times$ \\
3 & \textbf{+195.23} & +389.77 & +591.06 & +156{,}415 & 2.0$\times$ \\
4 & \textbf{+501.59} & +633.19 & +1{,}224 & +5{,}959{,}385 & 1.3$\times$ \\
\bottomrule
\end{tabular}
\end{table}

\subsection{Pythia-1B (16L, 1B Parameters)}

Pythia-1B has a nearly flat $\alpha$ profile ($\alpha \in [0.168, 0.229]$, $\Delta\alpha{=}0.061$) with the minimum at L12 ($\alpha{=}0.168$). Zone-aware spectral pruning targets these low-$\alpha$ interior layers and consistently outperforms Last-N (Table~\ref{tab:pythia_pruning}).

At $k{=}4$, zone-aware achieves $\Delta\text{PPL}{=}+55.32$ vs.\ Last-N $\Delta{=}+199.12$ (\textbf{3.6$\times$ better}), while spectral-worst (removing highest-$\alpha$ layers L3, L7, L8, L11) causes $\Delta{=}+15{,}146$ (\textbf{274$\times$ worse})---dramatic confirmation that $\alpha$ captures layer importance even in near-flat profiles.

\begin{table}[ht]
\centering
\caption{Pythia pruning ($\Delta$PPL). Zone-aware spectral outperforms Last-N on Pythia-1B (16L) and Pythia-160M (12L), and at aggressive pruning on Pythia-410M (24L). Baseline PPL: 1B = 12.41, 410M = 14.97.}
\label{tab:pythia_pruning}
\small
\begin{tabular}{llcccccc}
\toprule
Model & $k$ & Zone-Aware & Last-N & Random & Worst & L-N/ZA \\
\midrule
\multirow{4}{*}{\shortstack{1B\\(16L)}}
& 1 & \textbf{+5.60} & +6.10 & +4.87 & +40.40 & 1.1$\times$ \\
& 2 & \textbf{+11.83} & +15.23 & +11.61 & +86.58 & 1.3$\times$ \\
& 3 & \textbf{+20.90} & +53.36 & +27.95 & +11{,}703 & 2.6$\times$ \\
& 4 & \textbf{+55.32} & +199.12 & +117.92 & +15{,}146 & 3.6$\times$ \\
\midrule
\multirow{6}{*}{\shortstack{410M\\(24L)}}
& 1 & +7.34 & \textbf{+4.40} & +4.43 & +4.40 & 0.6$\times$ \\
& 2 & +23.77 & \textbf{+10.12} & +11.11 & +7.00 & 0.4$\times$ \\
& 4 & +83.20 & \textbf{+41.89} & +121.21 & +33.08 & 0.5$\times$ \\
& 6 & \textbf{+258.24} & +310.01 & +320.82 & +99.23 & 1.2$\times$ \\
& 8 & \textbf{+707.73} & +1{,}383.35 & +2{,}129 & +29{,}548 & 2.0$\times$ \\
\bottomrule
\end{tabular}
\end{table}

\subsection{Pythia-410M (24L, 410M Parameters)}

Pythia-410M has a strongly rising $\alpha$ profile ($\alpha \in [0.202, 0.522]$, $\Delta\alpha{=}0.320$) with a sharp phase transition at L14. This means the high-$\alpha$ layers are at the end (L17--L23)---exactly where Last-N removes layers. At small $k$, Last-N \emph{accidentally} targets low-importance layers (L20--L22 have high $\alpha$ but are near the output boundary). Zone-aware targets L1--L6 (low-$\alpha$ interior), which are foundational despite low $\alpha$.

However, at \textbf{aggressive pruning} ($k \geq 6$), the pattern reverses: zone-aware outperforms Last-N by $1.2$--$2.0\times$, because Last-N is forced to remove the critical high-$\alpha$ late layers that drive Pythia-410M's performance.

\begin{figure}[ht]
    \centering
    \includegraphics[width=\linewidth]{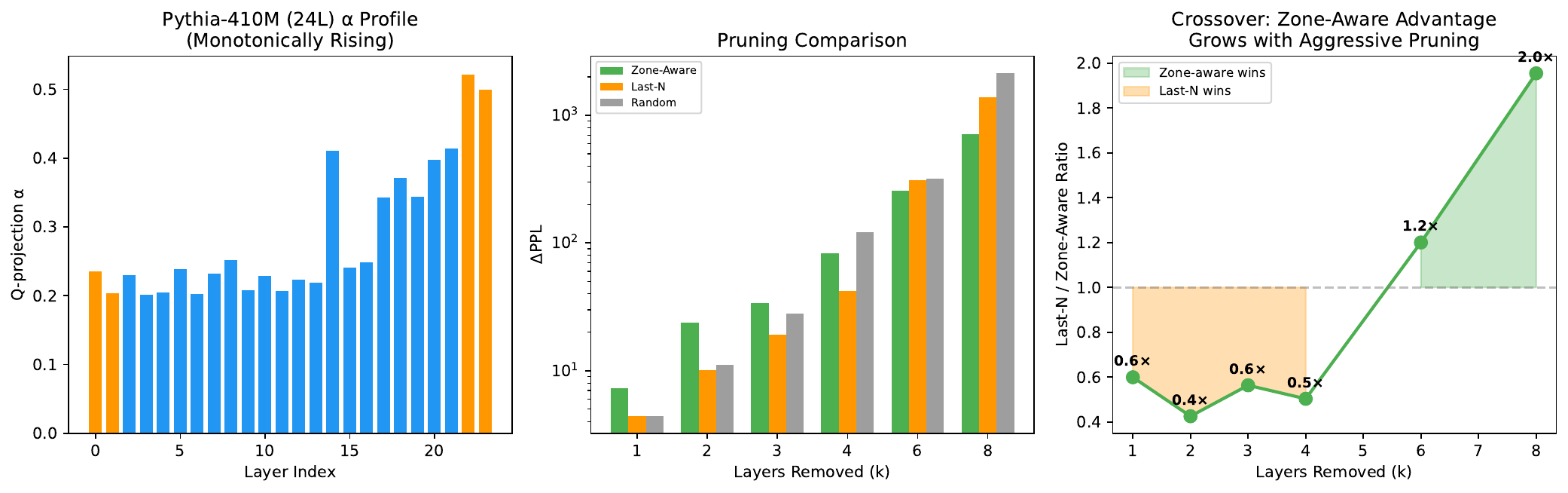}
    \caption{Pythia-410M (24L) pruning with crossover. (a)~Monotonically rising $\alpha$ profile. (b)~Zone-aware vs.\ Last-N on log scale. (c)~Ratio showing crossover at $k{=}6$: zone-aware dominates at aggressive pruning.}
    \label{fig:pythia410m_pruning_app}
\end{figure}

\subsection{Topology Dependence: A Key Insight}

The pruning results across seven models in two families reveal a fundamental insight: the optimal pruning strategy depends on $\alpha$ \emph{topology}. When the $\alpha$ peak is early (GPT-2 after long training), low-$\alpha$ layers concentrate in the second half, and spectral pruning outperforms Last-N. When $\alpha$ rises monotonically (Pythia-160M/410M/1B), low-$\alpha$ layers are foundational early layers, and naive removal is catastrophic---but zone-aware pruning restores the advantage (up to $2.0\times$ for Pythia-160M, $3.6\times$ for Pythia-1B). Pythia-410M's crossover effect further confirms topology dependence: the advantage appears only at aggressive pruning ratios. This topology dependence \emph{strengthens} our claim that $\alpha$ encodes genuine structural information: it is not merely a proxy for layer position, but captures the specific computational role each layer plays, which varies systematically across training regimes and architectures.

\section{D16 Dynamics: Additional Figures}
\label{app:d16}

This appendix collects supplementary figures for the D8/D12/D16 custom-training experiments and should be read as the high-resolution companion to the main-text dynamics section. The purpose is not to introduce a new claim, but to make the temporal structure of the training trajectory visually explicit at a granularity that would be too expensive to include in the main paper. In particular, these figures show how stable-rank collapse, $\alpha$ divergence, matrix-type specialization, and pruning sensitivity emerge together rather than as isolated observations.

\begin{figure}[ht]
    \centering
    \includegraphics[width=\linewidth]{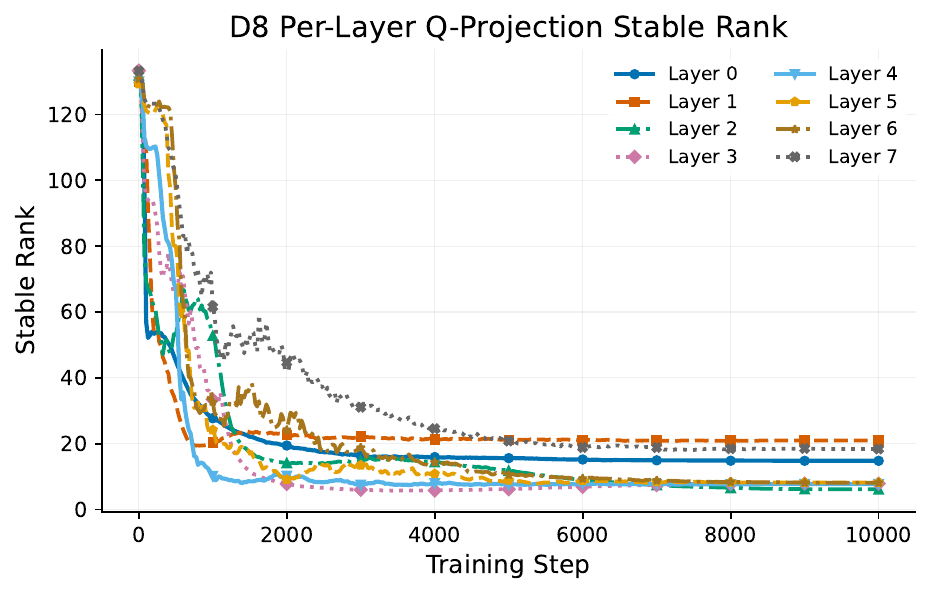}
    \caption{D8 per-layer Q-projection SR trajectories. Each color is one layer.}
    \label{fig:d8_sr_app}
\end{figure}

\begin{figure}[ht]
    \centering
    \includegraphics[width=\linewidth]{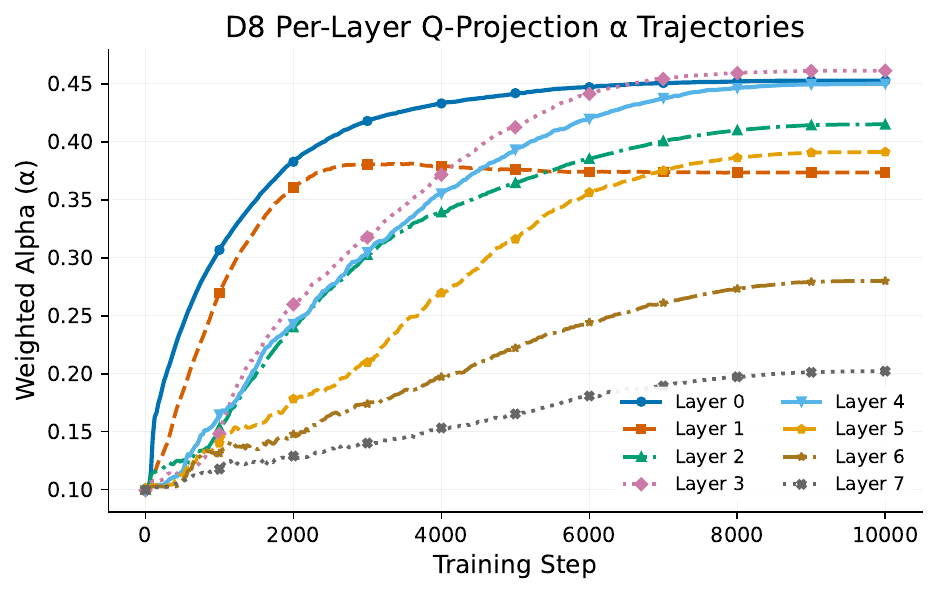}
    \caption{D8 per-layer Q-projection $\alpha$ trajectories. Layers diverge early and never reconverge.}
    \label{fig:d8_alpha_app}
\end{figure}

\begin{figure}[ht]
    \centering
    \includegraphics[width=\linewidth]{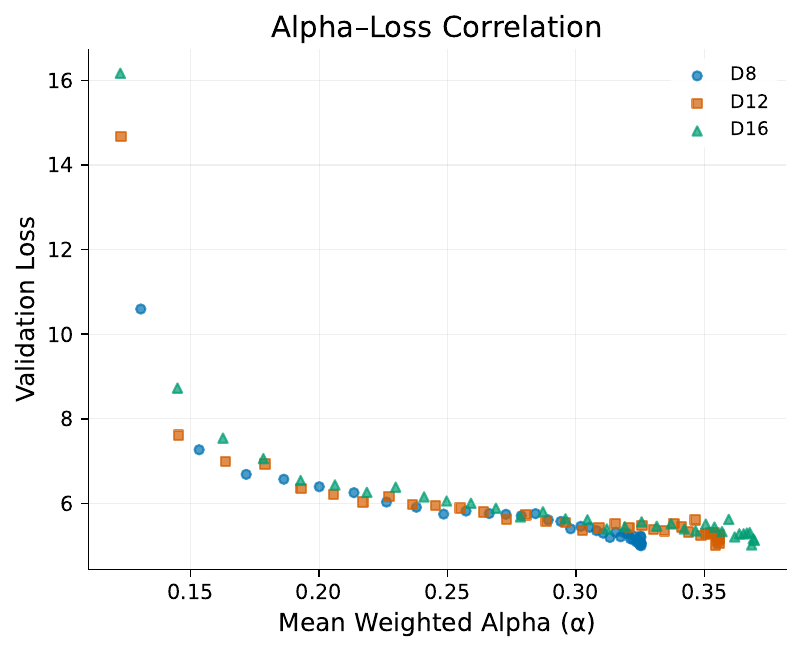}
    \caption{$\alpha$--loss correlation across models.}
    \label{fig:alpha_loss_app}
\end{figure}

The first pair of plots (Figures~\ref{fig:d8_sr_app} and~\ref{fig:d8_alpha_app}) provides the most direct intuition for the two-timescale story. Stable rank drops rapidly and relatively uniformly at the beginning of training, whereas the per-layer $\alpha$ trajectories separate early and remain separated. This is the appendix-level visualization behind the claim that compression equilibrates while spectral-shape specialization persists. Figure~\ref{fig:alpha_loss_app} then connects this geometric separation back to optimization by showing that models with larger $\alpha$ separation also tend to exhibit clearer loss-linked spectral structure.

\begin{figure}[ht]
    \centering
    \includegraphics[width=\linewidth]{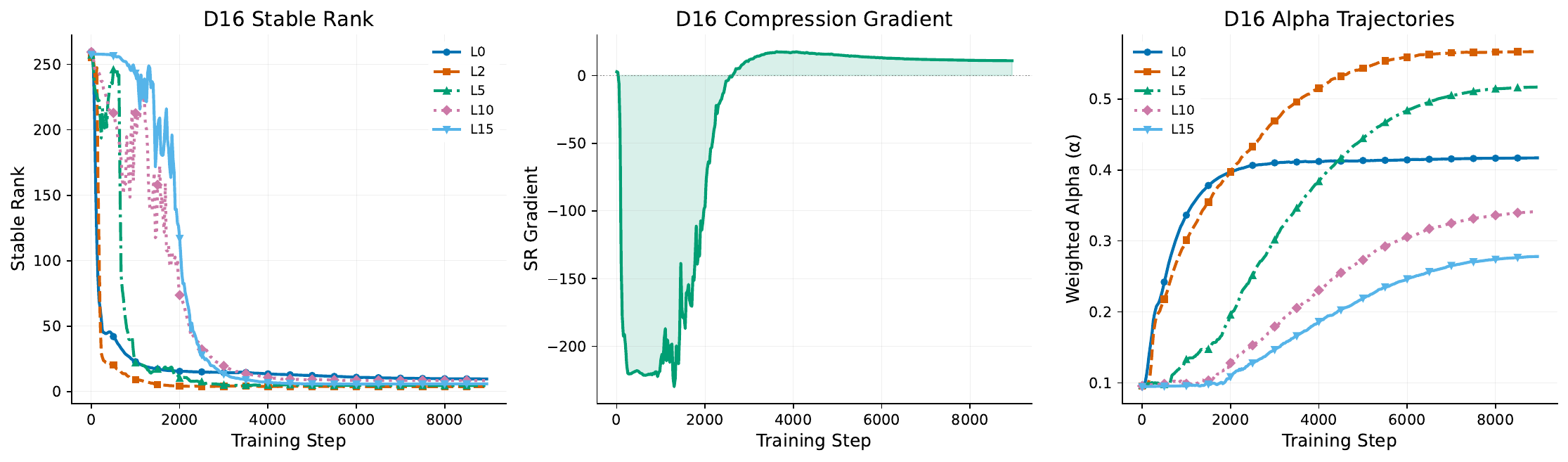}
    \caption{D16 three-panel dynamics.}
    \label{fig:d16_dynamics_app}
\end{figure}

\begin{figure}[ht]
    \centering
    \includegraphics[width=\linewidth]{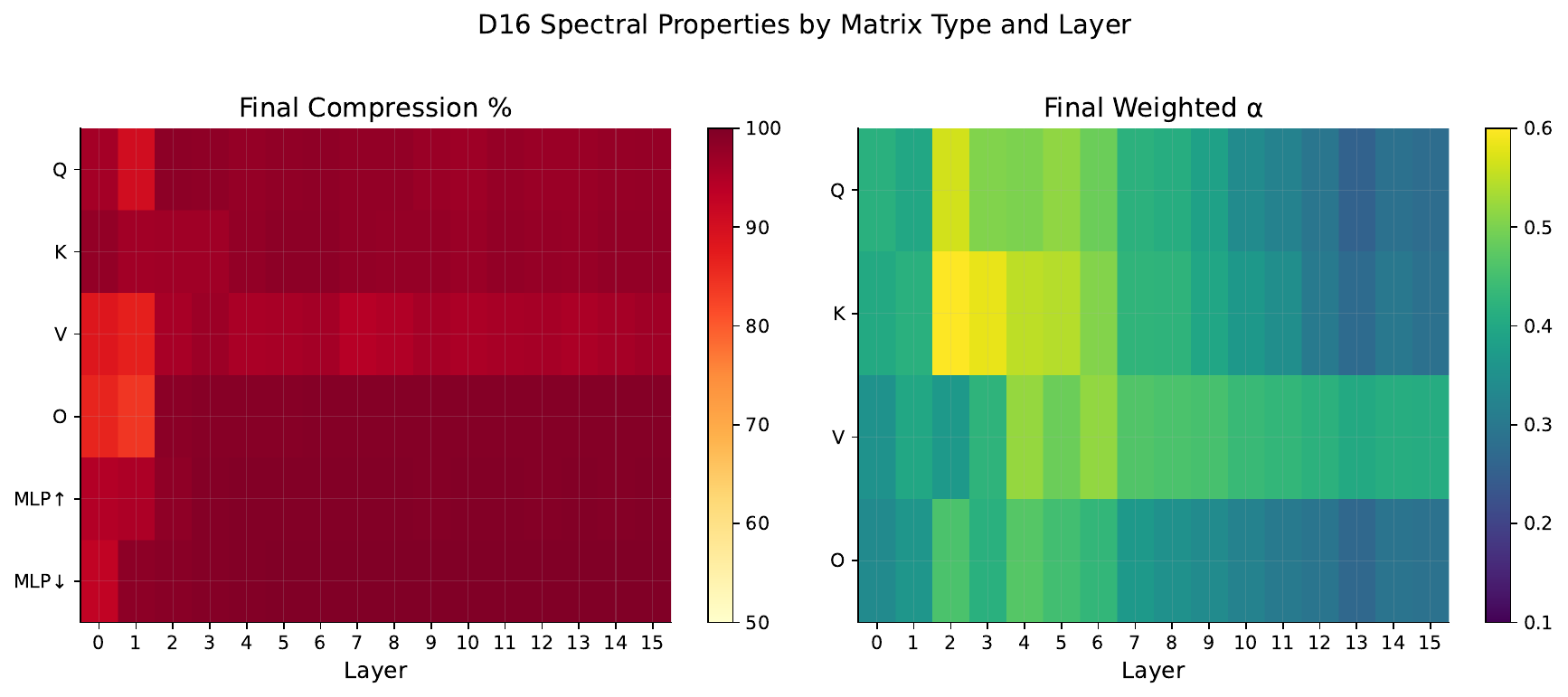}
    \caption{D16 matrix type heatmap.}
    \label{fig:d16_heatmap_app}
\end{figure}

\begin{figure}[ht]
    \centering
    \includegraphics[width=0.48\linewidth]{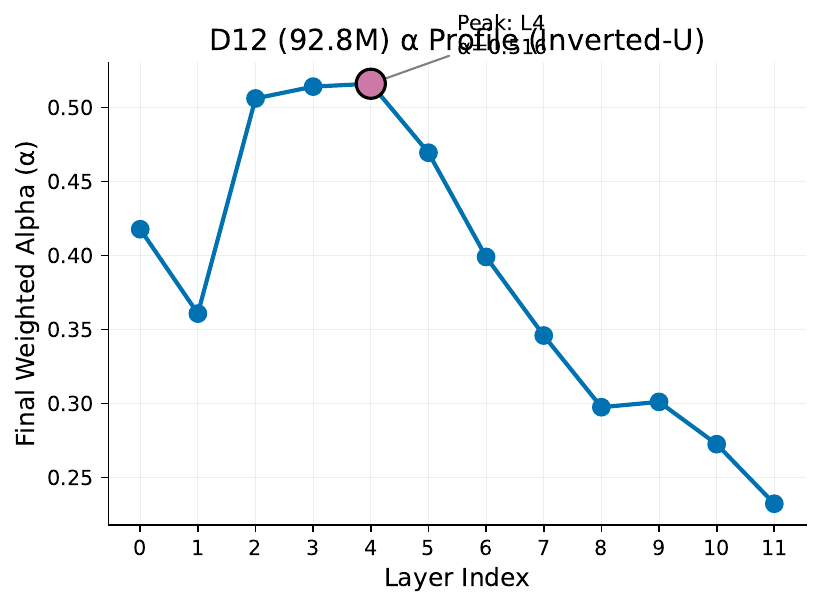}
    \caption{D12 $\alpha$ profile (inverted-U).}
    \label{fig:d12_alpha_app}
\end{figure}

Figures~\ref{fig:d16_dynamics_app} and~\ref{fig:d16_heatmap_app} zoom in on the 16-layer model, which is the clearest custom example of a wide, structured depth profile. The three-panel dynamics figure makes it clear that D16 is not merely a scaled-up D8: the deeper network develops a broader middle-layer band of elevated $\alpha$, and the matrix-type heatmap shows that this band is driven primarily by Q/K-like specialization rather than uniform change across all weights. Figure~\ref{fig:d12_alpha_app} is included as an intermediate checkpoint showing that D12 already exhibits the inverted-U topology, which then sharpens further in D16.

\begin{figure}[ht]
    \centering
    \includegraphics[width=\linewidth]{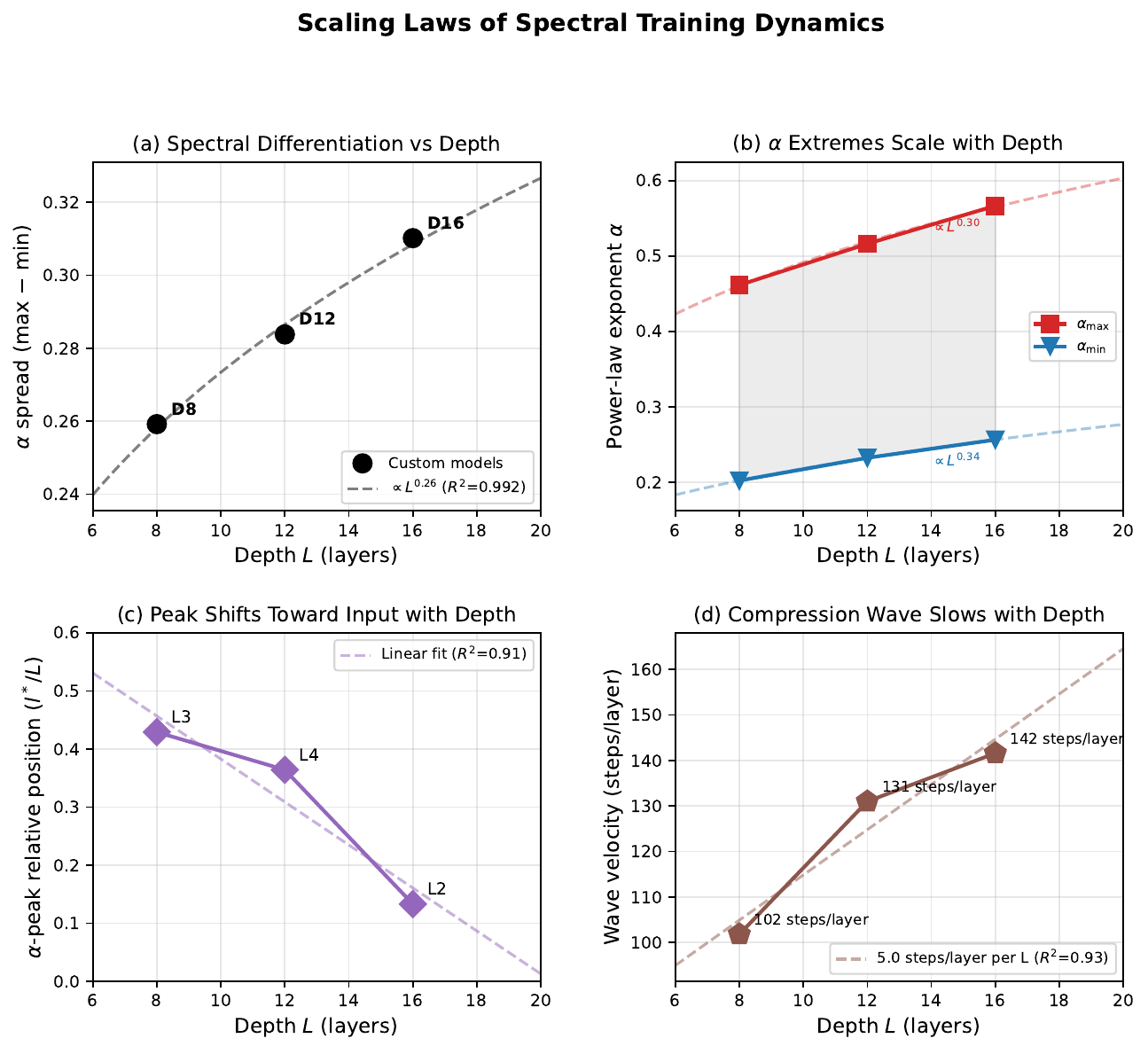}
    \caption{Scaling laws: (a)~$\Delta\alpha$ vs.\ depth, (b)~$\alpha$ extremes, (c)~peak position, (d)~wave velocity.}
    \label{fig:scaling_laws_app}
\end{figure}

\begin{figure}[ht]
    \centering
    \includegraphics[width=\linewidth]{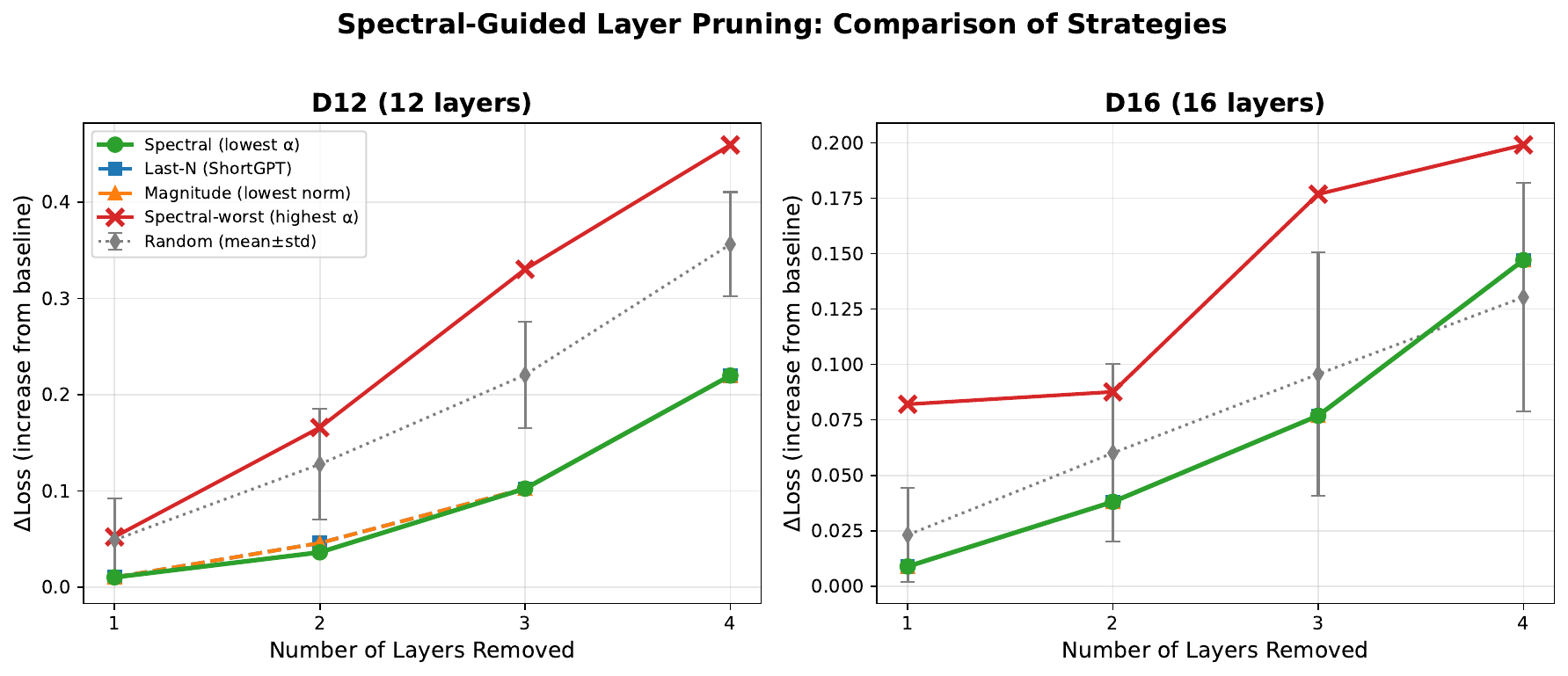}
    \caption{Pruning strategy comparison (line plots).}
    \label{fig:pruning_app}
\end{figure}

\begin{figure}[ht]
    \centering
    \includegraphics[width=\linewidth]{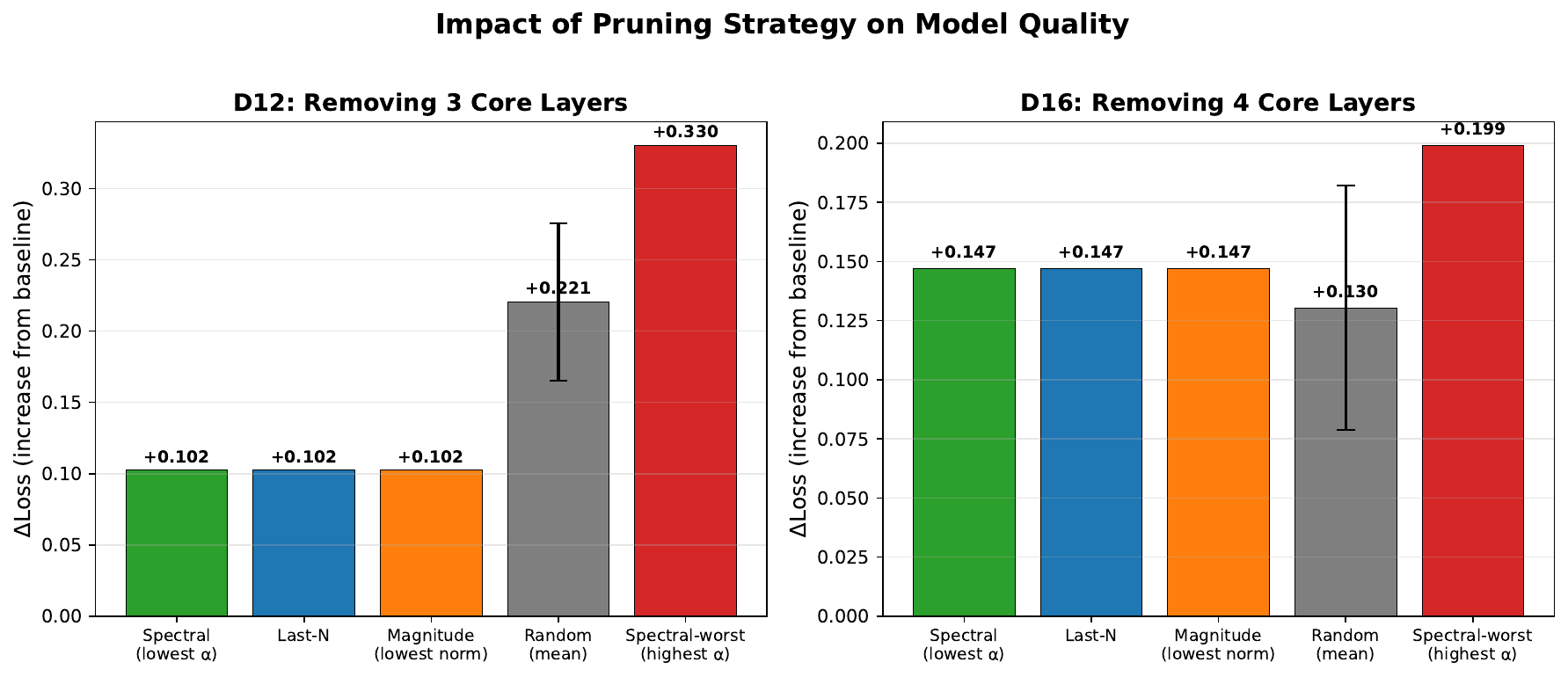}
    \caption{Pruning strategy comparison (bar charts).}
    \label{fig:pruning_bars_app}
\end{figure}

The final set of figures links the dynamics story to downstream utility. Figure~\ref{fig:scaling_laws_app} summarizes the empirical scaling relations that emerge from the custom family, while Figures~\ref{fig:pruning_app} and~\ref{fig:pruning_bars_app} show that these geometric patterns are operationally meaningful for intervention: low-$\alpha$ layers can be pruned with systematically smaller degradation than naive alternatives. In other words, the D16 supplementary figures close the loop from training dynamics to functional consequence.

\end{document}